\documentclass[11pt]{article}
\usepackage[final]{acl}             

\usepackage{times}
\usepackage{latexsym}
\usepackage[T1]{fontenc}
\usepackage[utf8]{inputenc}
\usepackage{microtype}
\usepackage{inconsolata}

\usepackage{amsmath,amssymb}
\usepackage{booktabs}
\usepackage{multirow}
\usepackage{graphicx}
\usepackage{hyperref}
\usepackage{xcolor}
\usepackage{url}

\usepackage{tikz}
\usetikzlibrary{positioning,arrows.meta,fit,backgrounds,shapes.geometric}

\usepackage{pgfplots}
\pgfplotsset{compat=1.18}
\usepgfplotslibrary{groupplots}

\providecommand{\citet}[1]{\cite{#1}}
\providecommand{\citep}[1]{\cite{#1}}
\providecommand{\citealt}[1]{\cite{#1}}

\title{The Illusion of Control: Why Bare Classifier Inversion Silently
       Fails\\ in Concept-Bottleneck Text Generation}

\author{
  Qi Bing \\
  Shanghai Jiao Tong University \\
  \texttt{peach19981011@gmail.com} \\\And
  Xiaowei Shao \\
  Shanghai Jiao Tong University \\
  \texttt{xw.shao@sjtu.edu.cn} \\}

\begin{document}
\maketitle

\begin{abstract}
Concept-bottleneck controllable generation routes multi-attribute
control through a low-dimensional concept code that, at deployment, must
be synthesised from a target attribute configuration. We study this
problem in concept-bottleneck text generation under multi-axis
compositional generalisation, comparing three ways to obtain the
inference-time code: classifier inversion against the encoder heads,
reference-text encoding, and a post-hoc label-conditioned prior.
Since a concept code admits no direct LM-fluency term, regularising
inversion must instead constrain the code toward the encoder's training
distribution. We therefore test bare inversion and three regularised
variants: label-agnostic and label-conditioned Mahalanobis penalties, and
a conditional normalising-flow density baseline. Every inversion variant we test underperforms a simple post-hoc prior
fitted to per-combination encoder means on the same checkpoints, across
three backbone families spanning $124$M to $8$B parameters. The bare form of classifier inversion also silently collapses to chance, traceable to a directly measured off-manifold code. We validate this diagnosis on real-world benchmarks and under external
evaluators, enabling fair comparison with published baselines.
\end{abstract}

\section{Introduction}
\label{sec:intro}

Multi-attribute controllable text generation (MCTG) requires generating
text that satisfies several attribute axes
$\mathbf{c} = (c_1, \dots, c_A)$, and is commonly evaluated through
compositional generalisation, where performance is measured by
accuracy on attribute combinations held out during training
\citep{keysers2020,zhong2024compmctg}. Recent
concept-bottleneck language models (CB-LLMs) \citep{sun2025cb} show that
language generation can be routed through interpretable concept units and
steered by intervening on them, making the CB-LLM design a natural
starting point for concept-bottleneck MCTG.
In such a setting, deployment requires synthesising a code
$\mathbf{z}^\star$ from a target configuration $\mathbf{c}^\star$.
Since the bottleneck exposes classifier heads over $\mathbf{z}$, a
natural inference path is to optimise $\mathbf{z}$ against those heads.
The broader CTG steering literature typically combines such classifier
gradients with fluency, prototype-distance, or manifold regularisation
\citep{dathathri2019pplm,gu2022lens,gu2023prior}. Unlike token-level or
decoding-time steering, however, optimisation over a concept code has no
direct LM-fluency term; any regularisation must act through the code
distribution itself. As a result, the role and failure mode of
classifier-based code optimisation are often hidden inside compound
inference objectives. We therefore isolate this component in the
concept-bottleneck setting and ask what happens when the target
attributes are read out directly through the bottleneck classifier heads.

%
\begin{figure}[t]
\centering
\includegraphics[width=\columnwidth]{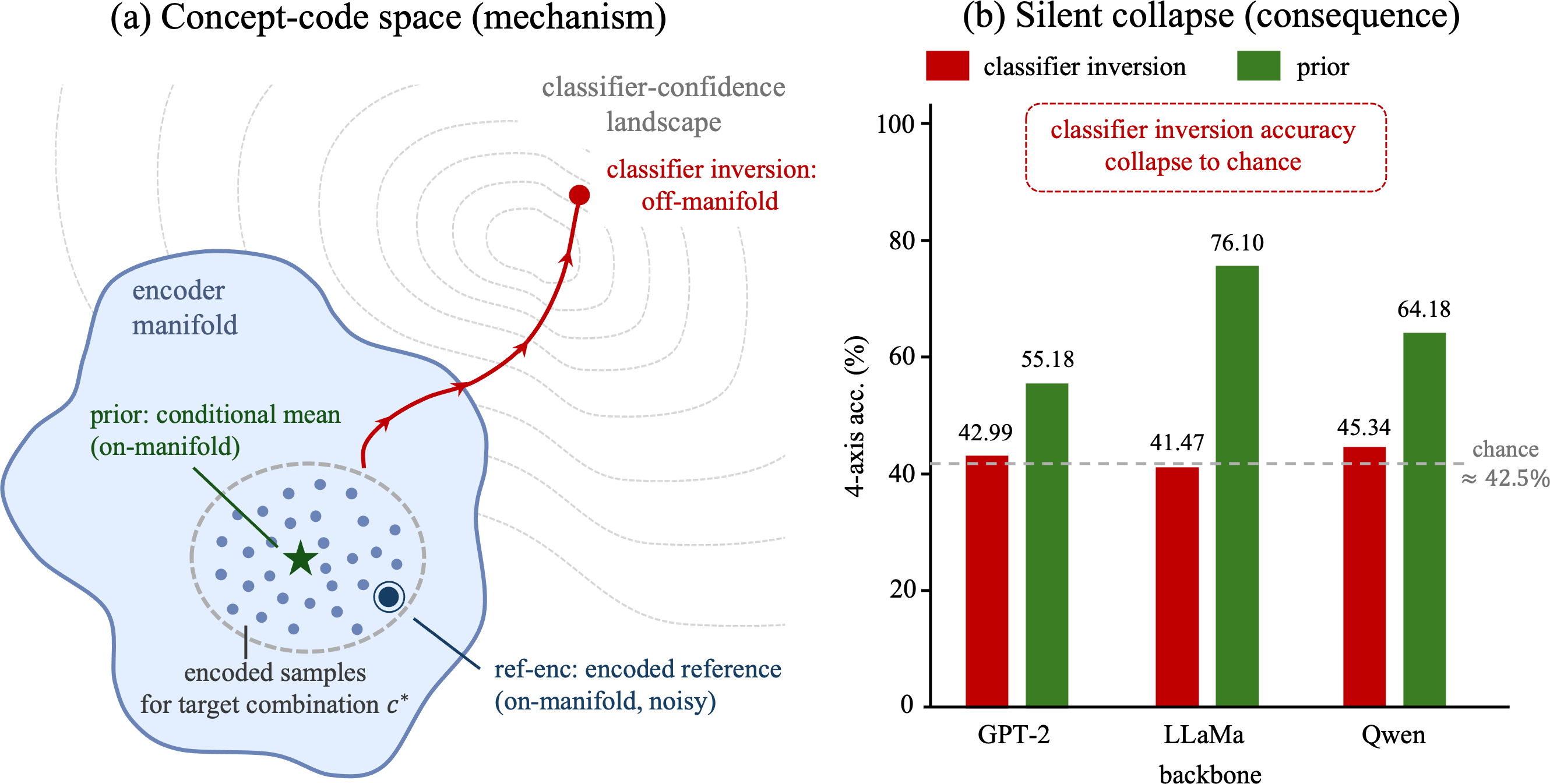}
\caption{\textbf{The illusion of control.} \textbf{(a)}~In concept-code
space, classifier inversion drifts off the encoder's training manifold;
a label-conditioned prior sits at the on-manifold per-combination mean,
and \texttt{ref-enc} is one on-manifold sample. \textbf{(b)}~CompMCTG
Fyelp 4-axis accuracy: \texttt{cls-inv} sits at the
$\approx\!42.5\%$ chance baseline while the prior recovers compositional
control on the \emph{same} checkpoints (split details:
Table~\ref{tab:fyelp}\,$^\dag$).}
\label{fig:teaser}
\end{figure}

We isolate this component experimentally
(Fig.~\ref{fig:teaser}) and identify a directly measured failure
mechanism: the inverted code lies $3$ to $7$ times farther from the
encoder's training distribution than a working code, both in
diagonal-Gaussian Mahalanobis distance and in mean distance to its $10$
nearest training neighbours. Adding a manifold regulariser to the
inversion objective lifts accuracy off chance (\S\ref{sec:modeb-fails}).
The corrective protocol is a single forward pass: freeze the trained encoder and fit a post-hoc MLP $g_\gamma$ to the per-combination means of the encoded training data. Given the target labels, the prior model estimates the conditional mean of the encoder code. Because the within-combination variation we measure exceeds the across-combination signal, the prior acts as a denoiser of sample-specific code variation, explaining its advantage over single-sample reference encoding (\S\ref{sec:analysis-prior}).

This paper makes three contributions.
\begin{itemize}
\item We systematically compare classifier inversion against a simple
post-hoc label-conditioned prior on matched checkpoints. We test bare
inversion and three regularised inversion variants (label-agnostic Mahalanobis, label-conditioned Mahalanobis, and a conditional normalising-flow density baseline) and find that every tested inversion variant underperforms the prior by $7$ to $29$\,pp (Tab.~\ref{tab:protocol-compare}; Apps.~\ref{app:clsinvreg},~\ref{app:flowprior}). The bare form
also collapses to chance, which we trace to a directly measured
off-manifold code (\S\ref{sec:modeb-fails}).
\item We introduce a deployable post-hoc label-conditioned prior that
recovers compositional generalisation on the same checkpoints, with no
inference-time optimisation and no reference text. The prior is fitted
on our proposed multi-axis concept-bottleneck architecture that extends
existing single-axis steering designs. To our knowledge, this is the
first concept-bottleneck text-generation protocol evaluated under
multi-axis compositional generalisation
(\S\ref{sec:prior-recovers},~\S\ref{sec:external}).
\item We show that the prior acts as a conditional-mean denoiser, and
provide evidence that the non-additive residual of the concept code is
dominated by per-sample encoder noise rather than measurable attribute
interaction (\S\ref{sec:additivity}).
\end{itemize}

\section{Related Work}
\label{sec:related-work}

%
%
\begin{figure*}[t]
\centering
\includegraphics[width=\textwidth]{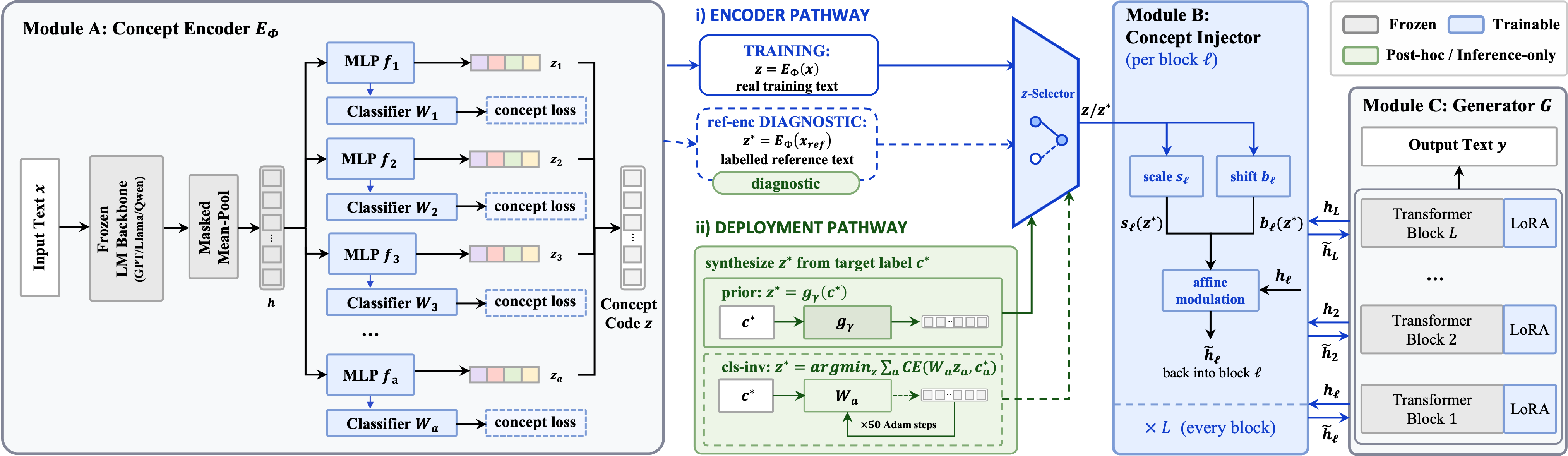}
\caption{Architecture of the concept-bottleneck CTG framework
(\S\ref{sec:method}). \textbf{Module A} encodes input text
$\mathbf{x}$ into a concatenated concept code
$\mathbf{z}=(\mathbf{z}_1,\dots,\mathbf{z}_A)$ using a frozen-LM encoder,
per-axis MLPs, and classifier heads. \textbf{z-Selector} supplies the
injector input: during training, $\mathbf{z}=E_\phi(\mathbf{x})$; during
inference, $\mathbf{z}^\star$ is obtained by one of the protocols in
\S\ref{sec:method-inference} (\texttt{prior}, \texttt{cls-inv}, or
\texttt{ref-enc}). \textbf{Module B} maps the selected code to per-layer
AdaLN-zero updates. \textbf{Module C} is a LoRA-adapted frozen
Transformer generator that emits $\mathbf{y}$ autoregressively.}
\label{fig:arch}
\end{figure*}

\paragraph{Concept-bottleneck language models.}
Concept-bottleneck models constrain output through a low-dimensional,
interpretable code, originating in vision \citep{koh2020concept} and
recently adapted to language as text-classification bottlenecks
\citep{tan2024tbm,sun2024c3m,bhan2025ctcbm,labadietamayo2025scbm,chaudhary2025essaycbm}
and, for generation, as CB-LLMs \citep{sun2025cb}, which evaluate
single-axis steering on each dataset. To our knowledge, this
paper is the first to extend the per-axis bottleneck from single-axis
steering to multi-axis compositional generalisation on combinations
unseen during training --- the setting in which the inference-protocol
pathology surfaces (\S\ref{sec:modeb-fails}).

\paragraph{Compositional and multi-attribute CTG.}
Multi-attribute CTG includes joint-training methods
\citep{keskar2019ctrl,yang2023catprompt,qian2022contrastive,zeng2023dcg},
decoding-time classifier-gradient or logit-biasing methods
\citep{dathathri2019pplm,yang2021fudge,krause2020gedi,liu2021dexperts},
latent-space control \citep{gu2022lens,gu2023prior}, adapter fusion
\citep{roy2024macs}, hidden-state edits \citep{kumar2023chrt},
token-level RL \citep{li2024tole}, and learning-free FFN reweighting
\citep{feng2024freectrl}. Closest to our prior is
\citet{gu2023prior}, who map an encoder posterior to a Gaussian via a
normalising flow. Instead, we fit a deterministic post-hoc
label-to-latent MLP without an invertibility constraint. Our generator is
conditioned through hidden-state injection, a mechanism related to FiLM
\citep{perez2018film}, AdaLN \citep{peebles2023dit}, prefix-tuning
\citep{li2021prefix}, and adapters \citep{houlsby2019adapter}, and
implemented with AdaLN-zero and rank-8 LoRA \citep{hu2022lora}. We evaluate on CompMCTG Fyelp \citep{zhong2024compmctg}, built from the Maximum Compound Divergence split \citep{keysers2020}, under its official RoBERTa-large evaluator. 

\paragraph{Inference protocols for concept-conditioned generation.}
Inference-time construction of the concept code remains under-specified
in concept-bottleneck generation. Existing classifier-gradient methods
such as PPLM \citep{dathathri2019pplm}, classifier-guided diffusion
\citep{li2022diffusionlm,liu2023latentops}, and activation steering
\citep{oozeer2025ksteering,li2023iti} optimise control representations
with additional fluency, prototype-distance, or distributional
regularisation \citep{dathathri2019pplm,gu2022lens,gu2023prior}. In
contrast, our focus is the inference protocol exposed by a trained
concept bottleneck. We compare several protocol choices on matched checkpoints and find that the post-hoc prior consistently dominates the tested inversion
variants. Other steering work constructs control vectors without classifier inversion, e.g., by mean-difference or contrastive construction
\citep{hsu2026clas,lee2024cast,stolfo2024instructsteer}.
\section{Background}
\label{sec:background}
Since our target is an inference protocol, we first fix the object of
study: a multi-axis concept-bottleneck architecture for compositional
CTG, extending the single-axis concept-bottleneck generation design of
\citet{sun2025cb}. This architecture poses the inference problem we
study: synthesising an inference-time concept code from a target
attribute configuration. It serves as the experimental vehicle for our protocol diagnosis.

\subsection{Framework}
\label{sec:method}

\paragraph{Problem formulation.}
A controllable text generation task specifies $A$ attribute axes, each
with a finite value set $\mathcal{A}_a$. A \emph{concept configuration}
is
$\mathbf{c} = (c_1, \dots, c_A) \in \mathcal{A} := \prod_a
\mathcal{A}_a$. Given $\mathbf{c}$, the model generates text
$\mathbf{y}$ that satisfies all specified attributes, modelling
$p_\theta(\mathbf{y} \mid \mathbf{c})$. During training, the generator is
teacher-forced on $\mathbf{x}$ in an autoencoding setup, with
$\mathbf{y} := \mathbf{x}$. Following the MCD protocol
\citep{keysers2020,zhong2024compmctg}, training covers only
$\mathcal{C}_{\text{seen}} \subset \mathcal{A}$; we report performance
on both seen and unseen combinations, where
$\mathcal{C}_{\text{unseen}} := \mathcal{A} \setminus
\mathcal{C}_{\text{seen}}$.

\paragraph{Architecture.} The architecture has three modules connected through continuous concept vectors (Fig.~\ref{fig:arch}).
First, a \emph{concept encoder}
$E_\phi : \mathcal{X} \to \mathbb{R}^{Ad_c}$ mean-pools a frozen
backbone representation and passes it through $A$ per-axis MLPs $f_a$,
producing sub-codes $\mathbf{z}_a \in \mathbb{R}^{d_c}$ with per-axis
classifier heads $\mathbf{W}_a$. The full concept code is
$\mathbf{z} = (\mathbf{z}_1,\dots,\mathbf{z}_A)$. Second, a
\emph{concept injector} $I_\psi$ maps the code to per-layer gated
residual updates for the generator blocks; our default implementation
uses AdaLN-zero \citep{peebles2023dit}. Third, the \emph{generator} $G$
is a frozen pretrained Transformer adapted with LoRA
\citep{hu2022lora}.
During training, the injector receives the encoder code
$\mathbf{z}=E_\phi(\mathbf{x})$ and maps it to per-layer updates
$\{\Delta\mathbf{h}_\ell\}$. At inference, the encoder path is bypassed, so an
inference-time code $\mathbf{z}^\star$ must be synthesised from the
target configuration $\mathbf{c}^\star$
(\S\ref{sec:method-inference}). Training uses a teacher-forced LM loss, a
per-axis concept-classification loss, and an inter-axis orthogonality
regulariser; the injector update and four-phase training schedule are
given in App.~\ref{app:arch}.

\subsection{Inference Protocols}
\label{sec:method-inference}

Any procedure that maps a target configuration $\mathbf{c}^\star$ to an
inference-time code $\mathbf{z}^\star$ defines a distinct evaluation
protocol. We compare three code sources.

\paragraph{Classifier inversion (\textsc{Cls-Inv}).}
Given the exposed classifier heads, the direct classifier-based protocol
optimises the code so that the heads predict the target labels:
\begin{equation}
\mathbf{z}^\star_{\text{cls}}
= \arg\min_{\mathbf{z}} \sum_a
\mathrm{CE}(\mathbf{W}_a \mathbf{z}_a, c_a^\star).
\label{eq:clsinv}
\end{equation}
The objective decouples over axes. We solve each axis with 50 Adam steps
from $\mathbf{z}_a^{(0)}=\mathbf{W}_a[c_a^\star,:]^\top$.

\paragraph{Reference-text encoding (\textsc{Ref-Enc}).}
As a single-sample diagnostic, we encode one labelled held-out example:
$\mathbf{z}^\star_{\text{ref-enc}} :=
E_\phi(\mathbf{x}_{\text{ref}})$, where
$\pi_a(\mathbf{x}_{\text{ref}}) = c_a^\star$. We do not treat this as an
upper bound. The encoded reference carries the surface idiosyncrasies of
one sentence, so a conditional-mean estimator can outperform it by
averaging away sample-specific variation. We include \textsc{Ref-Enc}
alongside \textsc{Prior} to make this variance source explicit
(\S\ref{sec:analysis-prior}); it is not deployable, since it requires a
labelled reference sentence matching the requested configuration.

\paragraph{ Amortised label prior (\textsc{Prior}).} After the main
model has been trained, we freeze $E_\phi$ and fit a
post-hoc MLP $g_\gamma : \mathcal{A} \to
\mathbb{R}^{Ad_c}$ by
\begin{equation}
\mathcal{L}(\gamma) = \mathbb{E}_{(\mathbf{x},\mathbf{c}) \sim
\mathcal{D}_{\text{train}}} \big\| g_\gamma(\mathbf{c}) -
E_\phi(\mathbf{x}) \big\|_2^2.
\label{eq:prior}
\end{equation}
At inference, we use $\mathbf{z}^\star_{\text{prior}} :=
g_\gamma(\mathbf{c}^\star)$. The MLP $g_\gamma$ has a single hidden layer
of $128$ units (GELU) and is fitted post-hoc in well under a minute. It is
trained only on $\mathcal{C}_{\text{seen}}$ but queried on
$\mathcal{C}_{\text{unseen}}$. The encoder's per-axis structure lets the learned label-to-code map provide a compositional estimate of the per-configuration encoder mean.

\section{Experimental Setup}
\label{sec:setup}

Our evaluation centres on CompMCTG Fyelp, a real multi-attribute review
benchmark with compositional splits that supports the main claims. We
also use a synthetic 4-axis task and single-axis YelpP as controlled
checks.

\paragraph{Datasets.}
Our primary setting is \textbf{Fyelp} (CompMCTG,
4-axis sentiment $\times$ gender $\times$ cuisine $\times$ tense;
$65$K/$1.5$K/$1.5$K/$1{,}750$
train/val/\texttt{test\_seen}/\texttt{test\_unseen}) under two
compositional splits: Hold-Out idx=$-0$ ($39$ seen and $1$ unseen
combination) and \textbf{ACD} (half of the $40$ combinations unseen;
\S\ref{sec:fyelp}). We also evaluate on \textbf{Amazon} (CompMCTG,
2-axis sentiment $\times$ topic; App.~\ref{app:amazon}), and use
two controlled checks: a \textbf{Synthetic 4-axis MCD} task ($4^4$
combinations, 200 seen / 56 unseen via the MCD protocol
\citep{keysers2020}; \S\ref{sec:modeb-fails}--\S\ref{sec:prior-recovers})
and \textbf{YelpP} (binary sentiment, single-axis CB-LLMs cross-check;
App.~\ref{app:yelpp}).

\paragraph{Backbones.} Tab.~\ref{tab:fyelp} reports four Fyelp backbones
spanning 124M--1.5B: GPT-2 124M, GPT-2-Medium 355M (the backbone every
CompMCTG baseline uses, and therefore our matched comparison in
\S\ref{sec:fyelp}), LLaMA-3.2 1B (primary scale), and Qwen-2.5 1.5B. The
first three use AdaLN-zero; Qwen-2.5 1.5B uses the additive injector
after AdaLN-zero proved unstable. The synthetic sanity check
(\S\ref{sec:modeb-fails}) adds Qwen-2.5 0.5B and LLaMA-3.2 3B, and a
LLaMA-3 8B YelpP checkpoint covers the single-axis check. All models are
LoRA-adapted on frozen weights (rank~8, $\alpha=16$), trained for $25$
epochs with AdamW and a cosine schedule, with concept dimension
$d_c=32$.

\paragraph{Evaluation.} For Fyelp, we use the official CompMCTG 4-axis RoBERTa-large classifier suite \citep{zhong2024compmctg}, invoked verbatim through a subprocess
wrapper to match the published baseline pipeline; perplexity is computed
with GPT2-large under the same pipeline. We never use the bottleneck
encoder's own classifier heads for evaluation. We report per-axis and
4-axis-mean accuracy on generated text, joint all-axes-correct accuracy,
and fluency metrics (perplexity and Dist-$n$); headline numbers are on
\texttt{test\_unseen}, except for \textsc{Cls-Inv} on Hold-Out, which we
report on \texttt{test\_seen} because the single held-out combination
admits a classifier-default artefact (App.~\ref{app:singleton}); the ACD
split gives the artefact-free within-split comparison.
Tab.~\ref{tab:fyelp} marks the affected cells. For comparability, we fix one validation-selected
stable recipe per backbone and a fixed evaluation seed $42$; the full
seven-point comparability policy is in App.~\ref{app:training}.

\paragraph{Artefacts.} Every benchmark, evaluator, and backbone we use is
a public third-party release under its own licence. We release our code
under the MIT Licence, together with the trained concept encoder,
classifier heads, injector, LoRA adapter, and label prior for all four
main-table backbones on both
splits.\footnote{\url{https://github.com/BiancaBing/cbctg-illusion-of-control}}

\section{Classifier inversion silently collapses across backbones}
\label{sec:modeb-fails}

Classifier inversion is the most direct classifier-based answer to the
inference problem in \S\ref{sec:method-inference}. We isolate this
procedure from the regularised objectives that usually surround
classifier-gradient methods (\S\ref{sec:related-work}) and find that it
silently fails: on every trained checkpoint, it produces text with little
or no concept signal while training metrics remain healthy. Through
direct measurement and regulariser ablations, we trace the collapse to a
consistent mechanism: classifier inversion drives the control code off
the encoder's training distribution.

\subsection{The collapse: chance-level accuracy on every backbone}
We applied classifier inversion (Eq.~\ref{eq:clsinv}) to every trained
checkpoint and evaluated the generated text under the official CompMCTG
Fyelp evaluator. On the seen split ($39$ combinations), which we
headline for \textsc{Cls-Inv} for the reason given in
\S\ref{sec:setup}, 4-axis accuracy is $41.47\%$ on LLaMA-3.2 1B,
$42.99\%$ on GPT-2 124M, $45.34\%$ on Qwen-2.5 1.5B, and $46.88\%$ on
GPT-2-Medium 355M. Every backbone is within $5$\,pp of the $42.5\%$
random baseline, and $12$ to $35$\,pp below the prior's
\texttt{test\_unseen} accuracy on the same checkpoints
(Fig.~\ref{fig:teaser}); the like-for-like ACD comparison in
\S\ref{sec:prior-recovers} shows the same ordering.
The generated text carries little concept
signal: it degenerates into token-repetition loops or coherent but
off-domain pretrain-mode text, and an LLM judge rates none of $20$
\textsc{Cls-Inv} samples as fluent reviews
(Apps.~\ref{app:samples},~\ref{app:judge}). \textbf{The collapse is
invisible from training metrics}: they remain healthy to the final epoch,
and the 50-step inversion optimiser \emph{does} maximise classifier
accuracy on $\mathbf{z}^\star$. The failure occurs at generation time.

\paragraph{The same collapse on a controlled sanity check.}
The collapse is not specific to Fyelp. On a synthetic 4-axis task with
marginal-independent attributes, \textsc{Cls-Inv} accuracy stays within
$\pm 0.03$ of the $0.25$ baseline on five backbones across three families
and 124M--3B parameters. On single-axis YelpP with LLaMA-3 8B, it remains
at the $0.50$ binary baseline
(App.~\ref{app:synthetic}, Tab.~\ref{tab:syn-collapse}). The failure
appears under both injector mechanisms --- AdaLN-zero and the additive
fallback for Qwen-1.5B --- and is total on the joint all-axes-correct
metric ($\le 0.005$). This points to the inference protocol rather than a
single backbone or injector design.

\subsection{The mechanism: an off-manifold control code}
\label{sec:analysis-modeb}

The classifier-inversion objective (Eq.~\ref{eq:clsinv}) contains no term
that keeps $\mathbf{z}^\star$ within the code distribution on which the
generator was trained. We show that the inverted code leaves this
distribution, over-drives the injector, and that constraining it back
toward the encoder manifold repairs the collapse.

\paragraph{The inverted code is measurably off-manifold.}
We fit a per-dimension diagonal Gaussian
$(\boldsymbol{\mu},\boldsymbol{\sigma})$ to the encoder codes of $4000$
training reviews, the code distribution on which the AdaLN injector was
trained, and measure how far each inference-time $\mathbf{z}^\star$ lies
from it (App.~\ref{app:offmanifold}). Classifier inversion is a clear
outlier on both backbones: its Mahalanobis distance is $3.3$--$3.7$,
compared with $0.50$--$0.60$ for the prior and $\approx 1.0$ for
\textsc{Ref-Enc}. This gives a $3$--$7\times$ gap, with the same
separation in nearest-neighbour distance. The prior and \textsc{Ref-Enc}
lie inside the encoder's code distribution; the inverted code does not.
This is measured directly, not inferred from downstream generation
quality.

\begin{figure}[t]
\centering
\begin{tikzpicture}
\begin{groupplot}[
  group style={
    group size=2 by 1,
    horizontal sep=16mm,
  },
  width=0.49\linewidth,
  height=4.9cm,
  xlabel={transformer block index $\ell$},
  ylabel near ticks,
  ylabel style={font=\small, yshift=-1pt},
  xlabel style={font=\small, yshift=2pt},
  title style={font=\small\bfseries, yshift=-4pt},
  tick label style={font=\scriptsize},
  legend style={font=\scriptsize, draw=gray!45, fill=white,
                fill opacity=0.92, text opacity=1,
                cells={anchor=west}, inner sep=2.5pt,
                row sep=-1pt, nodes={inner sep=1pt}},
  every axis plot/.append style={line width=1.0pt},
  grid=both,
  grid style={line width=.1pt, draw=gray!18},
  major grid style={line width=.2pt, draw=gray!38},
  enlarge x limits=0.04,
]

\nextgroupplot[
  title={LLaMA-3.2 1B Fyelp},
  ylabel={per-layer rel\_mod (log)},
  ymode=log,
  ymin=0.4, ymax=160,
  xmin=0, xmax=15,
  xtick={0,3,6,9,12,15},
  ytick={1,10,100},
  yticklabels={1,10,100},
  legend pos=north west,
  legend cell align={left},
]
\addplot[color=blue!65!black, mark=*, mark size=1.0, mark options={fill=blue!65!black}]
coordinates {
  (0,0.6043) (1,0.7018) (2,0.7046) (3,0.7268)
  (4,0.7558) (5,0.7766) (6,0.7960) (7,0.8132)
  (8,0.8097) (9,0.7978) (10,0.7975) (11,0.8041)
  (12,0.8313) (13,0.8727) (14,0.9054) (15,0.9702)
};
\addlegendentry{\texttt{ref-enc}}

\addplot[color=red!72!black, mark=square*, mark size=1.0, mark options={fill=red!72!black}]
coordinates {
  (0,3.293) (1,3.705) (2,3.815) (3,4.823)
  (4,6.476) (5,8.038) (6,11.742) (7,15.609)
  (8,25.423) (9,32.909) (10,42.318) (11,49.809)
  (12,61.742) (13,65.916) (14,82.743) (15,84.963)
};
\addlegendentry{\texttt{cls-inv}}

\nextgroupplot[
  title={GPT-2-124M Fyelp},
  ylabel={per-layer rel\_mod (linear)},
  ymin=0.0, ymax=5.4,
  xmin=0, xmax=11,
  xtick={0,2,4,6,8,10},
  ytick={0,1,2,3,4,5},
  legend pos=north west,
  legend cell align={left},
]
\addplot[color=blue!65!black, mark=*, mark size=1.0, mark options={fill=blue!65!black}]
coordinates {
  (0,0.4414) (1,0.4280) (2,0.3506) (3,0.3387)
  (4,0.3844) (5,0.4060) (6,0.4253) (7,0.4542)
  (8,0.4776) (9,0.4976) (10,0.4962) (11,1.1107)
};
\addlegendentry{\texttt{ref-enc}}

\addplot[color=red!72!black, mark=square*, mark size=1.0, mark options={fill=red!72!black}]
coordinates {
  (0,0.4779) (1,0.5089) (2,0.6842) (3,0.8620)
  (4,0.9060) (5,0.9197) (6,1.1368) (7,1.2535)
  (8,1.4279) (9,1.4836) (10,1.4813) (11,4.5942)
};
\addlegendentry{\texttt{cls-inv}}

\end{groupplot}
\end{tikzpicture}

\caption{Activation-level signature of classifier inversion
(\S\ref{sec:analysis-modeb}). We plot per-layer relative modulation
$\mathrm{rel\_mod}_\ell =
\|\tilde{\mathbf{h}}_\ell-\mathbf{h}_\ell\|/\|\mathbf{h}_\ell\|$,
averaged over tokens and prompts, on trained Fyelp checkpoints.
\textsc{Ref-Enc} remains bounded on both backbones, while
\textsc{Cls-Inv} produces depth-amplified over-modulation: severe on
LLaMA-3.2 1B (log $y$-axis) and milder but still visible on GPT-2 124M.
The corresponding off-manifold distances are reported in
Table~\ref{tab:offmanifold}.}
\label{fig:modeab}
\end{figure}
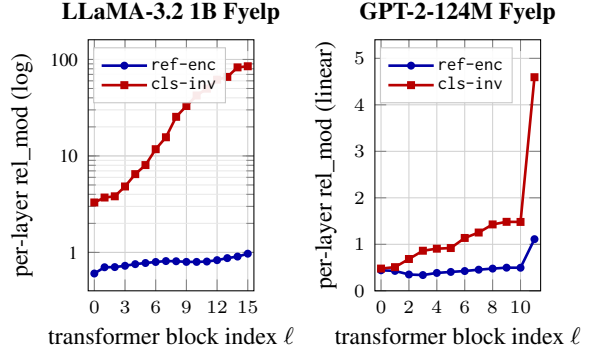

\paragraph{Off-manifold codes over-drive the injector.}
An off-manifold $\mathbf{z}^\star$ pushes the AdaLN injector outside its
trained range. We probe per-layer relative modulation,
$\mathrm{rel\_mod}_\ell =
\|\tilde{\mathbf{h}}_\ell-\mathbf{h}_\ell\|/\|\mathbf{h}_\ell\|$, on the
trained Fyelp checkpoints (Fig.~\ref{fig:modeab}). On LLaMA-3.2 1B,
classifier inversion produces a depth-amplified perturbation: $\mathrm{rel\_mod}$
rises from $3.3$ at layer~0 to $85$ at layer~15, averaging a
$\mathbf{40\times}$ over-shoot over \textsc{Ref-Enc}. On GPT-2 124M, the
same over-shoot is present but milder ($2.7\times$). Both backbones show
the same signature: an off-manifold code amplified through depth, with a
magnitude that depends on the backbone family. The resulting text
degenerates into high-perplexity pretrain-mode loops (LLaMA-1B
perplexity $>130$). 

%
%
%
%

\paragraph{Does a regulariser close the gap to the prior?}
A standard manifold regulariser repairs the bare-form collapse, and more
targeted regularisers narrow the gap further. However, no tested variant
matches the post-hoc prior. We re-run inversion with a label-agnostic
manifold regulariser on the per-axis objective,
$\mathrm{CE}(\mathbf{W}_a\mathbf{z}_a, c_a^\star) +
\beta R(\mathbf{z}_a)$, sweeping $\beta$ under two penalty forms, two
backbones, and three seeds (App.~\ref{app:clsinvreg}). A
\emph{Mahalanobis} penalty,
$R =
\mathbb{E}[((\mathbf{z}_a-\boldsymbol{\mu}_a)/\boldsymbol{\sigma}_a)^2]$,
lifts 4-axis accuracy from chance to $57.4\%$ on GPT-2 and $52.8\%$ on
LLaMA, while reducing perplexity from $\sim\!130$ to $\sim\!20$. A
\emph{shell} penalty is zero inside the $\pm2\sigma$ ellipsoid and has no
centring pull, avoiding a trivial mean-collapse explanation. It still
lifts accuracy by $+8.5$ and $+9.0$\,pp; every lift exceeds seed noise
($\le 1.6$\,pp) by roughly an order of magnitude.

\begin{table*}[t]
\centering
\small
\begin{tabular}{llcc}
\toprule
Method / z-source & Backbone & \textbf{HO} & \textbf{ACD} \\
\midrule
\multicolumn{4}{l}{\textit{CompMCTG-paper baselines, GPT-2-Medium 355M (quoted from \citealt{zhong2024compmctg})}} \\
PPLM \citep{dathathri2019pplm}         & GPT-2-M 355M & 42.49 & 40.63 \\
Fudge \citep{yang2021fudge}            & GPT-2-M 355M & 41.71 & 40.42 \\
CatPrompt \citep{yang2023catprompt}    & GPT-2-M 355M & 64.57 & 54.77 \\
DCG \citep{zeng2023dcg}                & GPT-2-M 355M & 66.39 & 64.71 \\
Con.Prefix \citep{qian2022contrastive} & GPT-2-M 355M & 67.50 & 63.93 \\
CTRL \citep{keskar2019ctrl}            & GPT-2-M 355M & 68.29 & 65.31 \\
Prior \citep{gu2023prior}              & GPT-2-M 355M & 54.64 & 52.29 \\
Dis-Lens \citep{gu2022lens}            & GPT-2-M 355M & 67.06 & 57.31 \\
Meta-CTRL \citep{zhong2024compmctg}    & GPT-2-M 355M & \textbf{68.69} & \textbf{65.77} \\
\midrule
\multicolumn{4}{l}{\textit{Ours --- 65K-train \texttt{\_full} runs, 4-axis mean accuracy (\%)}} \\
\textsc{Cls-Inv}$^\dag$        & GPT-2 124M    & 42.99 & 40.14 \\
\textsc{Ref-Enc}                & GPT-2 124M    & 47.64 & 49.46 \\
\textbf{\textsc{Prior}}        & GPT-2 124M    & \textbf{55.18} & \textbf{55.91} \\
\textsc{Cls-Inv}$^\dag$        & GPT-2-M 355M  & 46.88 & 43.72 \\
\textsc{Ref-Enc}                & GPT-2-M 355M  & $49.79_{\pm0.08}$ & $46.73_{\pm0.06}$ \\
\textbf{\textsc{Prior}}        & GPT-2-M 355M  & $\mathbf{61.50}_{\pm0.67}$ & $\mathbf{59.52}_{\pm0.32}$ \\
\textsc{Cls-Inv}$^\dag$        & LLaMA-3.2 1B  & $41.47_{\pm0.46}$ & 34.51 \\
\textsc{Ref-Enc}                & LLaMA-3.2 1B  & $63.80_{\pm3.16}$ & 67.79 \\
\textbf{\textsc{Prior}}        & LLaMA-3.2 1B  & $\mathbf{76.10}_{\pm0.06}$ & \textbf{69.53} \\
\textsc{Cls-Inv}$^\dag$        & Qwen-2.5 1.5B & 45.34 & 34.82 \\
\textsc{Ref-Enc}                & Qwen-2.5 1.5B & 60.90 & 62.04 \\
\textbf{\textsc{Prior}}        & Qwen-2.5 1.5B & \textbf{64.18} & \textbf{60.48} \\
\midrule
\multicolumn{4}{l}{\textit{Within-split sanity, ACD \texttt{test\_unseen}$^\ddag$ (artefact-free; App.~\ref{app:acdclsinv} for per-axis)}} \\
\textsc{Cls-Inv}$^\ddag$ & GPT-2 124M / LLaMA-1B / Qwen-1.5B & --- & 48.47\,/\,50.75\,/\,52.82 \\
\textsc{Ref-Enc}                   & GPT-2 124M / LLaMA-1B / Qwen-1.5B & --- & 49.46\,/\,67.79\,/\,62.04 \\
\textbf{\textsc{Prior}}            & GPT-2 124M / LLaMA-1B / Qwen-1.5B & --- & \textbf{55.91\,/\,69.53\,/\,60.48} \\
\bottomrule
\end{tabular}
\caption{4-axis mean accuracy (\%) on CompMCTG Fyelp under the official
RoBERTa-large evaluator \citep{zhong2024compmctg}; Hold-Out (HO) and
ACD splits, idx=$-0$, chance baseline $\approx 42.5\%$. Unmarked rows
report \texttt{test\_unseen} (the compositional-generalisation number);
$^\dag$\,\textsc{Cls-Inv} on \texttt{test\_seen} (its Hold-Out
\texttt{test\_unseen} number is inflated by a singleton artefact,
App.~\ref{app:singleton}); $^\ddag$\,artefact-free
\texttt{test\_unseen} on the ACD split, whose half-combination hold-out
admits no singleton --- the corresponding within-split lift over the
\textsc{Prior} cells is $+7.4$ to $+18.8$\,pp. Multi-seed entries
(42/7/11) are marked $_{\pm\text{sd}}$. \textbf{Backbones differ}: every
CompMCTG baseline uses GPT-2-Medium 355M; the strictly matched
comparison is our GPT-2-Medium 355M block, where \textsc{Prior} does
\emph{not} surpass Meta-CTRL on either split; the LLaMA-1B win is
cross-scale ($\approx 3.5\times$ the baseline backbone). Per-axis
breakdown in Tab.~\ref{tab:fyelp-peraxis} (App.~\ref{app:peraxis}).}
\label{tab:fyelp}
\end{table*}

A manifold constraint, under either penalty and on either backbone,
converts a chance-level protocol into a working one, supporting the
off-manifold diagnosis. This is the failure mode that regularisation in
classifier-gradient methods helps prevent.
The best regularised inversion variant still trails the label prior by
$7$ to $29$\,pp on \texttt{test\_seen}, as does a conditional
normalising-flow density baseline
(Tab.~\ref{tab:protocol-compare};
Apps.~\ref{app:clsinvreg},~\ref{app:flowprior},~\ref{app:clsinvgrid}
for an inversion-side hyperparameter grid). We therefore adopt the prior
as the recommended protocol.


\section{A label-conditioned prior recovers compositional control}
\label{sec:prior-recovers}

If classifier inversion is an unreliable way to obtain a control code from
a target configuration, what alternative inference protocol should be
used? We show that a post-hoc label-conditioned prior, which estimates
the conditional mean of the encoder code given the target labels,
recovers compositional control on the same checkpoints where classifier
inversion fails, with no inference-time optimisation and no reference
text.

\subsection{Prior inference recovers compositional generalisation}

We evaluate the post-hoc label-conditioned MLP $g_\gamma$ on CompMCTG Fyelp under the official RoBERTa-large 4-axis evaluator. We compare it against two reference protocols: the
collapsed classifier-inversion protocol \textsc{Cls-Inv} and the single-sample reference encoding \textsc{Ref-Enc}.

\paragraph{Prior inference recovers control on every backbone.} While classifier inversion remains near the $\approx 42.5\%$ random
baseline even on \texttt{test\_seen}
(Tab.~\ref{tab:fyelp}\,($^\dag$)), the prior reaches $55.18\%$
(GPT-2 124M), $61.50\%$ (GPT-2-Medium 355M), $64.18\%$
(Qwen-2.5 1.5B), and $76.10\%$ (LLaMA-3.2 1B) 4-axis accuracy on the
Hold-Out \texttt{test\_unseen} split --- a $+12$ to $+35$\,pp recovery on
identical checkpoints. The comparison is confirmed like-for-like on ACD,
where half of all attribute combinations are unseen and both protocols can
be read off the same split without the singleton artefact: on ACD
\texttt{test\_unseen}, the prior improves over \textsc{Cls-Inv} by
$+7.4$ to $+18.8$\,pp
(Tab.~\ref{tab:fyelp}, within-split block;
App.~\ref{app:acdclsinv}).
The gain is not merely over the bare protocol. Tab.~\ref{tab:protocol-compare}
compares the prior with regularised inversion variants on matched
checkpoints. Both a label-agnostic Mahalanobis penalty
(App.~\ref{app:clsinvreg}) and a sharper label-conditioned variant,
which pulls $\mathbf{z}_a$ toward the per-(axis, target-label) marginal
mean, still trail the prior by $7$ to $29$\,pp. They also require a
50-step inner optimisation per query and a backbone-specific $\beta$
sweep, both of which the deterministic prior avoids.

\begin{table}[!t]
\centering\small
\setlength{\tabcolsep}{4pt}
\begin{tabular}{lcc}
\toprule
Protocol & GPT-2-124M & LLaMA-1B \\
\midrule
\textsc{Cls-Inv} bare                                  & $42.99$ & $41.47$ \\
$\;$+ Mahal. (agn.)\textsuperscript{$\diamond$}        & $57.52$ & $52.91$ \\
$\;$+ Mahal. (cond.)\textsuperscript{$\diamond$}       & $60.74$ & $52.73$ \\
\texttt{flow-prior} (cond.\ NF)\textsuperscript{$\dag$} & $57.21$ & $69.95$ \\
\textbf{\textsc{Prior}} (\,$g_\gamma$\,)               & $\mathbf{67.7}$ & $\mathbf{81.7}$ \\
\bottomrule
\end{tabular}
\caption{Inversion-protocol comparison on Fyelp Hold-Out
\texttt{test\_seen}, 4-axis mean accuracy (\%); the same checkpoints
report $55.18\%/76.10\%$ on \texttt{test\_unseen}
(Tab.~\ref{tab:fyelp}). \emph{agn.}/\emph{cond.}:
label-agnostic / label-conditioned per-axis manifold penalty
(App.~\ref{app:clsinvreg}; best
$\beta\in\{0.1,0.3,1.0,3.0\}$, \textsuperscript{$\diamond$}; the
label-agnostic row is the single-seed full \texttt{test\_seen} number,
matching the $57.4/52.8$ three-seed subset means in
Tab.~\ref{tab:clsinv-reg}).
\texttt{flow-prior}: conditional normalising flow $p(\mathbf{z}\mid c)$
(\textsuperscript{$\dag$}; App.~\ref{app:flowprior}); trails the MLP by $10.5$/$11.8$\,pp,
matching the denoiser prediction (\S\ref{sec:analysis-prior}).
Label-conditioned row uses the $312$-row subset
(Tab.~\ref{tab:clsinv-reg-cond}); others use full
\texttt{test\_seen}. Same ordering on ACD
(Tab.~\ref{tab:fyelp} within-split block).}
\label{tab:protocol-compare}
\end{table}

\paragraph{The prior also exceeds the encoded reference \textsc{Ref-Enc}.} On Fyelp Hold-Out, the prior improves over \textsc{Ref-Enc} on every
backbone by $+3.3$ to $+12.3$\,pp in Tab.~\ref{tab:fyelp}. The only exception
is Qwen-1.5B on ACD, where the prior is $-1.6$\,pp below
\textsc{Ref-Enc}. We do \emph{not} interpret this as surpassing an upper
bound: \textsc{Ref-Enc} is a single-sample diagnostic and carries the
surface idiosyncrasies of one reference sentence. A conditional-mean
estimator can therefore outperform it by averaging away sample-specific
variation, confirming that $g_\gamma$ acts as the intended
per-combination denoiser (\S\ref{sec:analysis-prior}). The same direction
holds on the synthetic check (App.~\ref{app:synthetic}), survives the
bf16/fp32 recipe shift within $\pm 1.0$\,pp (App.~\ref{app:dtype}), and is reproduced by an LLM judge on both
attribute match and fluency (App.~\ref{app:judge}). The prior also
beats a non-parametric nearest-seen-combination retrieval baseline on
every unseen split by $+2.6$ to $+9.5$\,pp
(App.~\ref{app:retrieval}).

\subsection{Mechanism: prior as a conditional-mean denoiser}
\label{sec:analysis-prior}

\paragraph{Why the conditional mean is the right target.}
Three properties of the two objectives account for the gap between them.
\emph{(i)} The inversion objective (Eq.~\ref{eq:clsinv}) depends on
$\mathbf{z}$ only through the per-axis classifier heads $\mathbf{W}_a$,
so nothing in it penalises displacement from the code distribution on
which the injector was trained; its minimisers are unconstrained on that
distribution. \emph{(ii)} The prior objective (Eq.~\ref{eq:prior}) is a
squared loss, so on the combinations it is fitted to its population
minimiser is the conditional mean,
\begin{equation}
g_\gamma(\mathbf{c}) =
\mathbb{E}\!\left[E_\phi(\mathbf{x}) \mid \mathbf{c}\right],
\qquad \mathbf{c} \in \mathcal{C}_{\text{seen}},
\label{eq:condmean}
\end{equation}
which averages away variation not shared across examples of the same
combination and which we measure to lie inside the encoder's code
distribution (\S\ref{sec:analysis-modeb}). \emph{(iii)} Whether that
conditional mean extends to combinations never seen in training depends
on how far the code factorises across axes, which we quantify below.

At training time, $E_\phi(\mathbf{x})$ encodes both per-combination
concept information and sentence-specific stylistic variation. Fitted to
the per-combination mean, $g_\gamma$ therefore acts as a denoised
prototype: it retains the shared code component of a combination and
discards idiosyncratic sentence-level variation. Consistent with this
view, prior PPL is lower than \textsc{Ref-Enc} PPL across all three
backbones.

\paragraph{The sample-specific noise is measurable and exceeds the
concept signal.} On the GPT-2-Medium checkpoint, the mean per-dimension
$\sigma$ of $E_\phi(\mathbf{x})$ within a fixed combination is $0.43$,
against only $0.33$ across the per-combination means: \textsc{Ref-Enc}
carries the full $0.43$ into the generator, whereas $g_\gamma$, the
conditional mean, discards it and retains the $0.33$ of concept signal.
A direct control, averaging $k$
same-combination reference encodings, matches the learned prior within
$0.4$\,pp at $k=256$ (App.~\ref{app:ksample}). Conversely, a
conditional normalising flow $p(\mathbf{z}\mid\mathbf{c})$ underperforms
the MLP by $10.5$/$11.8$\,pp, consistent with the view that preserving
within-combination variance is harmful for this inference protocol
(App.~\ref{app:flowprior}).
A synthetic concept-swap diagnostic shows
a compatible family-level pattern, with partial per-axis fidelity on
LLaMA and stronger entanglement on GPT-2/Qwen; we report it only as a
supplementary check (App.~\ref{app:swap}).

\paragraph{Additivity of the concept code, and why the prior generalises.}
\label{sec:additivity}
The prior can generalise to unseen combinations when the frozen encoder
code $\mathbf{z}$ is sufficiently structured by attribute axes. An
additive per-axis model
$\boldsymbol{\mu} + \sum_a \Delta_a(y_a)$ explains only
$31.7$--$47.6\%$ of the variance of $\mathbf{z}$, increasing with
backbone scale. However, the remaining non-additive residual is not
explained by measurable label co-occurrence: all six Fyelp axis pairs
have zero mutual information and Cram\'er's $V$, since CompMCTG samples
combinations uniformly and the axes are independent by construction. We
therefore interpret the residual primarily as a label-invariant encoder
artefact dominated by per-sample variation, the same variation discarded
by the conditional-mean prior. An additive label prior is therefore well
matched to Fyelp, and adding an explicit interaction term gives no
reliable gain on either backbone tested
(App.~\ref{app:additivity}).

\section{Baseline comparison and cross-benchmark robustness}
\label{sec:external}

The prior-over-classifier-inversion result of \S\ref{sec:prior-recovers} is already evaluated under the official CompMCTG RoBERTa-large evaluator (\S\ref{sec:setup}). We now compare it against published CompMCTG baselines on Fyelp, test cross-dataset robustness on a second CompMCTG benchmark, Amazon, and add a single-axis YelpP check against CB-LLMs under their independently built RoBERTa evaluator.

\subsection{Comparison with the CompMCTG baselines}
\label{sec:fyelp}

We place the prior result on CompMCTG Fyelp
(Tab.~\ref{tab:fyelp}) against the CompMCTG-paper baselines
\citep{zhong2024compmctg} under the same official evaluator. 

\paragraph{Matched-backbone comparison: GPT-2-Medium 355M.}
The only strictly matched comparison retrains our framework at
GPT-2-Medium 355M. In this setting, the prior reaches
$61.50\%$ on Hold-Out and $59.52\%$ on ACD in
4-axis unseen accuracy over three training seeds, slightly below the strongest baseline. However, the prior sits above the mean
of the nine published CompMCTG baselines on both splits
($60.15\%$ HO, $56.13\%$ ACD), and above the median on ACD, where it
beats PPLM, Fudge, Gu et al.'s Prior, CatPrompt, and Dis-Lens.
\textbf{The contribution is an inference-protocol correction, not an
accuracy record.} The matched setting nonetheless establishes the
within-method ordering with seed-level error bars far smaller than the
reported gaps: the prior beats \textsc{Ref-Enc} by $+12.8$\,pp on ACD
unseen and collapsed classifier inversion by $+14$--$16$\,pp, while
achieving roughly $3\times$ lower perplexity than Meta-CTRL. The 4-axis
score is depressed roughly uniformly by the near-chance gender axis; the
matched prior reaches $72.34\%$ on 3-axis ACD unseen.

\paragraph{Cross-scale result and scaling trend.} For completeness, on Fyelp Hold-Out our LLaMA-3.2 1B prior reaches
$76.10\%$ 4-axis unseen accuracy, $+7.41$\,pp over Meta-CTRL. We label
this as a cross-scale comparison rather than a state-of-the-art claim,
since LLaMA-3.2 1B has roughly $3.5\times$ the parameters of the
GPT-2-Medium baselines. Prior Hold-Out accuracy increases with backbone
size across our four backbones, a family-confounded trend showing only
that the protocol does not saturate in the tested range. Only the
LLaMA-1B point clears Meta-CTRL, and only cross-scale.

\paragraph{Second real-world dataset: Amazon.} 
To check that the result is not Fyelp-specific, we run the same
GPT-2-124M pipeline on CompMCTG \textbf{Amazon}, a product-review dataset
with two axes, sentiment $\times$ topic, under the official 2-axis
classifier suite (App.~\ref{app:amazon}). The ordering reproduces: on the
seen split, \textsc{Prior} reaches $76.5\%$, beating collapsed classifier
inversion by $+30.0$\,pp and \textsc{Ref-Enc} by $+14.4$\,pp, and it is
also best on the unseen split. The collapse is clearest on the multi-class
topic axis: classifier inversion is near chance ($20.2\%$ vs. $16.7\%$),
whereas the prior reaches $74.3\%$, with a $5\times$ perplexity gap. The
binary sentiment axis partly survives, but the content-bearing topic axis
collapses cleanly and the prior-over-\textsc{Ref-Enc} advantage
reproduces.

\paragraph{Single-axis check at a matched backbone.}
A minimal single-axis YelpP experiment against CB-LLMs
\citep{sun2025cb} uses LLaMA-3 8B backbone and RoBERTa evaluator
(App.~\ref{app:yelpp}). The like-for-like result is consistent with the
multi-axis findings: classifier inversion collapses to the $0.500$
binary-chance level, while \textsc{Prior} reaches $0.964$ steerability,
$+1.4$\,pp above the $0.95$ reported by CB-LLMs under our byte-identical
reproduction.

\section{Conclusion}
\label{sec:discussion}

Classifier inversion silently collapses to chance on every tested
backbone while training metrics remain healthy. We trace the failure to
an off-manifold inference-time code that over-drives the injector, and
confirm the diagnosis with a manifold-regulariser ablation. A post-hoc
label-conditioned prior, fitted to per-combination encoder means,
recovers compositional generalisation on the same checkpoints under
external evaluators and on a second real-world dataset. At a matched
backbone, it does not beat the strongest CompMCTG baseline; this is an
inference-protocol correction, not a state-of-the-art claim.
\textbf{For this family of concept-bottleneck CTG models, prior inference
should be the default $\mathbf{z}$-source rather than bare classifier
inversion.}


\section*{Limitations}
\label{sec:limitations}

\paragraph{Benchmark granularity.}
The Fyelp Hold-Out idx=$-0$ split holds out a single 4-axis combination,
so its \texttt{test\_unseen} score is a coarse single-combination
measurement and can inflate classifier-default artefacts
(App.~\ref{app:singleton}). We therefore also report ACD, where half of
all combinations are unseen, as the more reliable compositional-shift
measurement.

\paragraph{Architectural scope.}
Our collapse claim applies to concept-bottleneck CTG models whose
generator is conditioned through an injector trained on the encoder's
code distribution. This covers the CB-LLM-style architecture studied
here, but not all possible controllable-generation mechanisms.

\paragraph{Mechanistic scope.}
The off-manifold diagnosis is supported by distance measurements,
activation diagnostics, and manifold-regulariser ablations. However, it
does not fully explain why the same off-manifold displacement yields
different activation magnitudes across backbone families. A more detailed
analysis of injector sensitivity, depth effects, and hidden-state scaling
is left for future work.

\paragraph{Scale, language, and regulariser coverage.}
Experiments are limited to English benchmarks and to backbones up to 3B
for multi-axis CTG, with an 8B single-axis check. Multilingual, long-form,
and larger-scale multi-axis generation remain untested. Some
backbone--dataset cells are single-seed due to compute cost, although the
multi-seed blocks show protocol gaps larger than seed variation. Finally,
we do not claim that no regularised inversion method can close the gap:
we test label-agnostic and label-conditioned Mahalanobis penalties, a
shell penalty, and a conditional normalising-flow density baseline, but
broader regularisation schemes remain future work.

\section*{Ethical considerations}

The Fyelp benchmark includes a binary \textit{gender} attribute inherited
from prior style-transfer datasets. We treat this label only as a
benchmark-specific writing-style marker, not as an identity attribute, and
do not advocate generating gendered content about identifiable people.
Gender is used only through the standard CompMCTG evaluator; we do not
train or deploy an additional gender classifier beyond the benchmark
protocol.

This work diagnoses inference protocols on public benchmarks and releases
no new dataset or deployed model. The label-conditioned prior improves
the reliability of controlling already-labelled attributes, but does not
expand the set of controllable attributes beyond those present in the
training data. Misuse risks are therefore tied to the underlying
controllable-generation setting rather than to a new capability introduced
here.

\section*{Acknowledgements}
We used AI assistants for language polishing, copy-editing, and minor coding/debugging assistance. All scientific claims, experimental design, analyses, and final text were reviewed and verified by the authors, who take full responsibility for the submission.

\bibliography{references}

\clearpage
\appendix

\section*{Appendix}

\section{Training details}
\label{app:arch}

This appendix specifies the injector update, training objective, and loss
schedule for the framework in \S\ref{sec:method}.

\paragraph{Concept injector.}
\label{app:method-injector}
Our default injector is AdaLN \citep{peebles2023dit}. At every
generator block $\ell$, we apply
\begin{equation}
\tilde{\mathbf{h}}_\ell^{(t)}
= (1 + \mathbf{s}_\ell(\mathbf{z})) \odot \mathbf{h}_\ell^{(t)}
+ \mathbf{b}_\ell(\mathbf{z}),
\label{eq:adaln}
\end{equation}
where the scale and shift networks
$\mathbf{s}_\ell,\mathbf{b}_\ell$ are zero-initialised. For Qwen-2.5
1.5B, AdaLN-zero was unstable in our runs, so we use an additive fallback:
\begin{equation}
\tilde{\mathbf{h}}_\ell^{(t)}
= \mathbf{h}_\ell^{(t)}
+ s\,\sigma(g_\ell)\,P(\mathbf{z}),
\end{equation}
with gate initialisation $g_{\mathrm{init}}=-3.0$ and front scale $s=1$.
We use a negative gate initialisation rather than reducing $s$, since
the former only bounds the initial perturbation, whereas the latter also
scales the gradient through the perturbation at every update step.

\paragraph{Training objective.}
With trainable parameters
$\theta = (\phi,\psi,\theta_{\mathrm{LoRA}})$, we minimise
\begin{equation}
\mathcal{L}
= \mathcal{L}_{\mathrm{gen}}
+ \lambda_{\mathrm{c}}(\tau)\mathcal{L}_{\mathrm{concept}}
+ \lambda_{\mathrm{o}}(\tau)\mathcal{L}_{\mathrm{orth}},
\label{eq:total}
\end{equation}
where $\mathcal{L}_{\mathrm{gen}}$ is teacher-forced language-model
cross-entropy using $\mathbf{z}=E_\phi(\mathbf{x})$,
$\mathcal{L}_{\mathrm{concept}}=\sum_a
\mathrm{CE}(\hat{\mathbf{p}}_a,c_a)$ supervises the encoder classifier
heads, and
\begin{equation}
\mathcal{L}_{\mathrm{orth}}
=
\mathbb{E}\!\left[
\sum_{a\ne b}
\left(
\frac{\mathbf{z}_a^\top \mathbf{z}_b}
{\|\mathbf{z}_a\|\,\|\mathbf{z}_b\|}
\right)^2
\right]
\end{equation}
penalises inter-axis cosine similarity. We use a four-phase schedule:
first warming up $\mathcal{L}_{\mathrm{gen}}$, then ramping
$\lambda_{\mathrm{c}}$ and $\lambda_{\mathrm{o}}$ in sequence. The
per-phase weights are listed in Tab.~\ref{tab:schedule}.

\paragraph{Why no intervention-consistency term.}
Earlier versions included an intervention-consistency loss: when the
sub-vector for axis $a$ is replaced, the classifier-head predictions for
axes $b\ne a$ should remain unchanged. In our per-axis bottleneck, this
term is identically zero. Each classifier head $\mathbf{W}_b$ reads only
its own sub-vector $\mathbf{z}_b$, so replacing $\mathbf{z}_a$ for
$a\ne b$ leaves $\mathbf{W}_b\mathbf{z}_b$ unchanged. The loss and its
gradient are therefore zero by construction. We consequently use only
the three-term objective in Eq.~\ref{eq:total}. The orthogonality
regulariser remains active throughout phase~4 with weight $0.1$ for all
main-table checkpoints. All remaining constants were selected during
preliminary validation on the synthetic 4-axis split and then held fixed
across backbones.


\section{Single-axis YelpP check vs.\ CB-LLMs}
\label{app:yelpp}

To verify that the protocol findings are not artefacts of the multi-axis
MCD setup, we run a deliberately minimal single-attribute experiment
matched to the closest published concept-bottleneck generation work,
CB-LLMs \citep{sun2025cb}. CB-LLMs report a YelpP steerability score of
$\mathbf{0.95}$ with a LLaMA-3 8B backbone under a RoBERTa-base classifier
fine-tuned on \texttt{yelp\_polarity} (their Table~5). We train our
framework with the same LLaMA-3 8B base backbone, an AdaLN injector at
every block, rank-8 LoRA, fp32-encoder mixed precision, and $25$ epochs
on the $6{,}000$-sample YelpP single-axis split. We also re-run the
\texttt{test\_steerability.py} script from CB-LLMs on the official
\texttt{cesun/cbllm-generation} checkpoint, exactly reproducing their
reported score of $\mathbf{0.950}$ ($\Delta=0.000$). The comparison is
therefore head-to-head in backbone, evaluator, and metric.

\begin{table*}[t]
\centering
\small
\begin{tabular}{lcc}
\toprule
z-source & DistilBERT-SST-2 & CB-LLMs RoBERTa \\
\midrule
\textsc{Ref-Enc}  & 0.846 / 0.850 & 0.895 / 0.891 \\
\textbf{\textsc{Prior}}   & \textbf{0.915 / 0.921} & \textbf{0.964 / 0.964} \\
\textsc{Cls-Inv} & 0.500 / 0.500 & 0.916 / 0.918 \\
\midrule
CB-LLMs (official ckpt, BOS) & --- & \textbf{0.950} \\
\bottomrule
\end{tabular}
\caption{LLaMA-3 8B YelpP single-axis check (seen / unseen). Our
\textsc{Prior} reaches $0.964$ under the CB-LLMs RoBERTa evaluator,
$+1.4$\,pp above the reproduced CB-LLMs score at matched backbone and
evaluator. YelpP is single-axis, so \texttt{test\_seen} and
\texttt{test\_unseen} are random subsamples of the same distribution and
agree within $\le 0.7$\,pp. The high CB-LLMs RoBERTa score for
\textsc{Cls-Inv} reflects a default-positive prediction artefact on
lexically degraded generations; the DistilBERT score and perplexity
diagnostics show the collapse.}
\label{tab:yelpp}
\end{table*}

\paragraph{Result.}
Our \textsc{Prior} reaches $0.964 / 0.964$
(\texttt{test\_seen} / \texttt{test\_unseen}, $n=6{,}000$ each) under the
CB-LLMs RoBERTa evaluator, $+1.4$\,pp above the reproduced CB-LLMs score
(Tab.~\ref{tab:yelpp}). Unlike the cross-scale CompMCTG comparison, this
check is strictly backbone-matched, using LLaMA-3 8B on both sides. We
therefore read the result as like-for-like corroboration of the protocol
finding, not as a new state-of-the-art claim. The within-method
\textsc{Prior}$-$\textsc{Ref-Enc} gap is $+6.9$\,pp, consistent with the
conditional-mean denoising mechanism of \S\ref{sec:analysis-prior}.

Classifier inversion collapses to exactly $0.500$ under our internal
DistilBERT-SST-2 evaluator. Under the CB-LLMs RoBERTa evaluator, it
instead scores $0.916$, but the per-class breakdown (negative $0.83$,
positive $1.00$) shows this to be a default-positive evaluator artefact
on lexically degraded generations. The collapse is further confirmed by
perplexity: \textsc{Cls-Inv} has PPL $432$, compared with $17$ for
\textsc{Ref-Enc}, a $25\times$ increase at the LLaMA-3 8B scale.

\section{Second real benchmark: CompMCTG Amazon}
\label{app:amazon}

To test whether the prior-over-classifier-inversion result is specific to
Fyelp, we run the same GPT-2-124M pipeline on a second CompMCTG dataset,
\textbf{Amazon}. Amazon is a product-review benchmark with two attribute
axes: sentiment with $2$ values and topic with $6$ values, giving $12$
combinations. We use Hold-Out idx=$-0$, where the held-out combination is
\texttt{positive\_clothing}. The training recipe, three z-sources, and
official CompMCTG evaluation protocol are unchanged from the Fyelp runs;
only the dataset and 2-axis classifier suite differ.

\begin{table}[h]
\centering
\small
\begin{tabular}{llccc}
\toprule
z-source & split & sent. & topic & 2-axis \\
\midrule
\textsc{Cls-Inv}$^\dag$ & seen   & 72.8 & 20.2 & 46.5 \\
\textsc{Ref-Enc}        & seen   & 71.6 & 52.7 & 62.1 \\
\textbf{\textsc{Prior}} & seen   & \textbf{78.7} & \textbf{74.3} & \textbf{76.5} \\
\midrule
\textsc{Cls-Inv}$^\dag$ & unseen & 99.4 & \phantom{0}9.7 & 54.6 \\
\textsc{Ref-Enc}        & unseen & 78.4 & 27.8 & 53.1 \\
\textbf{\textsc{Prior}} & unseen & \textbf{94.9} & \textbf{61.2} & \textbf{78.1} \\
\bottomrule
\end{tabular}
\caption{CompMCTG Amazon results with GPT-2-124M under Hold-Out
idx=$-0$, evaluated by the official 2-axis Amazon classifier suite. We
report per-axis and 2-axis-mean accuracy (\%). Topic is six-way
classification, with $16.7\%$ chance accuracy; sentiment is binary, with
$50\%$ chance accuracy. \textsc{Prior} is the best protocol on both
splits. Classifier inversion collapses most clearly on the topic axis and
has a $5\times$ perplexity inflation on the seen split (PPL $65.1$ vs.
$13.0$ for \textsc{Prior}). $^\dag$\,On the unseen split, the held-out
combination is a singleton with positive sentiment; the high
\textsc{Cls-Inv} sentiment score is therefore a classifier-default
artefact (App.~\ref{app:singleton}).}
\label{tab:amazon}
\end{table}

\paragraph{The protocol ordering reproduces.}
As shown in Tab.~\ref{tab:amazon}, \textsc{Prior} is the best protocol on
both seen and unseen splits. On the seen split, it reaches $76.5\%$
2-axis accuracy, improving over classifier inversion by $+30.0$\,pp and
over \textsc{Ref-Enc} by $+14.4$\,pp. The latter gap reproduces the
conditional-mean denoising advantage of \S\ref{sec:analysis-prior} on a
second real benchmark and a new domain.

\paragraph{The collapse concentrates on the content-bearing axis.}
Classifier inversion is near chance on the six-way topic axis:
$20.2\%$ on seen and $9.7\%$ on unseen, compared with
$74.3\%$ and $61.2\%$ for \textsc{Prior}. The binary sentiment signal
partly survives in degraded text, so the 2-axis mean collapse is less
complete than the 4-axis Fyelp collapse. Nevertheless, the content-bearing
topic axis collapses cleanly, and the elevated classifier-inversion
perplexity ($65.1$ vs. $13.0$) matches the off-manifold signature
documented on Fyelp (\S\ref{sec:modeb-fails}).

\section{Synthetic 4-axis sanity check}
\label{app:synthetic}

The synthetic 4-axis MCD task (\S\ref{sec:setup}) has
marginally independent attributes and a controlled compositional split.
We use it only as a sanity check for the two main protocol findings:
classifier inversion collapses to the random baseline
(\S\ref{sec:modeb-fails}), and the prior improves over single-sample
reference encoding (\S\ref{sec:prior-recovers}). The synthetic task is
not load-bearing for the main natural-language claims, which are
supported by Fyelp, Amazon, and YelpP.

\begin{table*}[t]
\centering\small
\begin{tabular}{llcc}
\toprule
Backbone & Scale & Task & \textsc{Cls-Inv} mean \\
\midrule
GPT-2     & 124M & synthetic 4-axis & \textbf{0.239} \\
Qwen-0.5B & 500M & synthetic 4-axis & \textbf{0.276} \\
Qwen-1.5B & 1.5B & synthetic 4-axis & \textbf{0.239} \\
LLaMA-1B  & 1B   & synthetic 4-axis & \textbf{0.280} \\
LLaMA-3B  & 3B   & synthetic 4-axis & \textbf{0.224} \\
\midrule
LLaMA-3 8B & 8B  & YelpP binary\textsuperscript{$\dag$} & \textbf{0.500} \\
\bottomrule
\end{tabular}
\caption{Classifier-inversion collapse on the synthetic 4-axis sanity
check and single-axis YelpP. \textsc{Cls-Inv} mean accuracy on generated
text stays within $\pm 0.03$ of the random baseline on every backbone.
\textsuperscript{$\dag$}YelpP is single-axis binary, with random baseline
$0.50$; the synthetic rows are four-axis, with random baseline $0.25$.}
\label{tab:syn-collapse}
\end{table*}

The prior-over-\textsc{Ref-Enc} direction also reproduces on the
synthetic task (Tab.~\ref{tab:syn-prior}): the prior improves over
\textsc{Ref-Enc} on every backbone, consistent with the denoising pattern
observed on natural data.

\begin{table*}[t]
\centering\small
\begin{tabular}{lcccc}
\toprule
Backbone & \textsc{Cls-Inv} & \textsc{Ref-Enc} & \textbf{\textsc{Prior}} & $\Delta$ pp \\
\midrule
GPT-2 124M       & 0.239 & 0.536 & \textbf{0.574} & \textbf{$+3.8$} \\
Qwen-0.5B        & 0.276 & 0.603 & \textbf{0.679} & \textbf{$+7.55$} \\
Qwen-1.5B$^\dag$ & 0.239 & 0.360 & \textbf{0.379} & \textbf{$+1.9$} \\
\textbf{LLaMA-1B} ($n{=}5$) & 0.280 & \textbf{0.791 $\pm$ 0.021} & \textbf{0.840 $\pm$ 0.026} & \textbf{$+4.86 \pm 0.96$} \\
LLaMA-3B         & 0.224 & 0.857 & \textbf{0.907} & \textbf{$+4.93$} \\
\bottomrule
\end{tabular}
\caption{Synthetic 4-axis unseen mean accuracy for classifier inversion,
reference encoding, and the post-hoc prior. The prior improves over
\textsc{Ref-Enc} on every backbone, matching the direction observed on
natural data (\S\ref{sec:prior-recovers}). $^\dag$ Qwen-1.5B uses the
additive injector with gate initialisation $-3.0$. The LLaMA-1B row
reports the 5-seed mean $\pm$ standard deviation.}
\label{tab:syn-prior}
\end{table*}

\section{Per-axis Fyelp Hold-Out breakdown}
\label{app:peraxis}

Tab.~\ref{tab:fyelp} reports the 4-axis mean; this appendix gives the
per-axis Hold-Out \texttt{test\_unseen} accuracy behind it in Tab.~\ref{tab:fyelp-peraxis}.

\begin{table*}[t]
\centering\small
\begin{tabular}{llcccc}
\toprule
Method / z-source & Backbone & Acc$_{\text{s}}$ & Acc$_{\text{g}}$ & Acc$_{\text{c}}$ & Acc$_{\text{t}}$ \\
\midrule
\multicolumn{6}{l}{\textit{CompMCTG-paper baselines (GPT-2-Medium 355M)}} \\
PPLM       & GPT-2-M & 49.96 & 50.02 & 19.93 & 50.06 \\
Fudge      & GPT-2-M & 49.61 & 48.80 & 20.91 & 47.50 \\
CatPrompt  & GPT-2-M & 83.82 & 54.07 & 56.04 & 64.36 \\
DCG        & GPT-2-M & 90.29 & 56.39 & 57.00 & 61.88 \\
Con.Prefix & GPT-2-M & 93.66 & 59.24 & 48.30 & 68.78 \\
CTRL       & GPT-2-M & 87.88 & 59.65 & 59.02 & 66.61 \\
Prior      & GPT-2-M & 63.56 & 50.79 & 43.58 & 60.62 \\
Dis-Lens   & GPT-2-M & 77.03 & 56.05 & 78.23 & 56.93 \\
\midrule
\multicolumn{6}{l}{\textit{Ours, Hold-Out idx=$-0$, 65K-train \texttt{\_full} runs, single seed}} \\
\textsc{Cls-Inv} & GPT-2 124M    & 50.6 & 41.2 & 19.2 & 61.0 \\
\textsc{Ref-Enc}  & GPT-2 124M    & 61.6 & 35.1 & 36.9 & 57.0 \\
\textsc{Prior}   & GPT-2 124M    & 66.3 & 39.1 & 52.1 & 63.2 \\
\textsc{Cls-Inv} & GPT-2-M 355M  & 55.4 & 63.6 & 19.8 & 48.7 \\
\textsc{Ref-Enc}  & GPT-2-M 355M  & 52.9 & 41.4 & 39.8 & 64.9 \\
\textsc{Prior}   & GPT-2-M 355M  & 80.9 & 41.3 & 53.2 & 66.9 \\
\textsc{Cls-Inv} & LLaMA-3.2 1B  & 46.2 & 48.3 & 23.5 & 46.6 \\
\textsc{Ref-Enc}  & LLaMA-3.2 1B  & 87.5 & 62.3 & 50.6 & 62.3 \\
\textsc{Prior}   & LLaMA-3.2 1B  & 98.1 & 67.6 & 64.9 & 75.0 \\
\textsc{Cls-Inv} & Qwen-2.5 1.5B & 59.0 & 54.5 & 20.6 & 47.3 \\
\textsc{Ref-Enc}  & Qwen-2.5 1.5B & 80.6 & 62.8 & 38.2 & 61.9 \\
\textsc{Prior}   & Qwen-2.5 1.5B & 83.8 & 60.2 & 50.9 & 61.9 \\
\bottomrule
\end{tabular}
\caption{Per-axis CompMCTG Fyelp Hold-Out idx=$-0$ \texttt{test\_unseen}
accuracy (sentiment / gender / cuisine / tense), official RoBERTa-large
evaluator; our rows are 65K-train \texttt{\_full} runs, single seed
(seed~42). Random baselines: s/g/t $=50.0$, c $=20.0$. Averaging the
per-axis cells recovers the 4-axis means of Tab.~\ref{tab:fyelp}, except
its LLaMA-3.2 1B Hold-Out column: that column is a 3-seed mean whereas
these per-axis rows are the single seed-42 run, so they differ --- by
$0.3$\,pp for \textsc{Prior} and \textsc{Cls-Inv} but $1.9$\,pp for
\textsc{Ref-Enc}, whose single-combination unseen split is the most
seed-sensitive. \textsc{Cls-Inv} rows are reported on \texttt{test\_seen}
(per Tab.~\ref{tab:fyelp}, footnote~$\dag$).
Gender is the weakest-controlled axis for almost every method --- near or
below the $50\%$ chance level for the baselines and for classifier
inversion --- consistent with gender being barely encoded in $\mathbf{z}$
(\S\ref{sec:additivity}).}
\label{tab:fyelp-peraxis}
\end{table*}

\section{Qualitative generation samples}
\label{app:samples}

Tab.~\ref{tab:samples} shows representative generations for one target
configuration on the GPT-2-124M Fyelp Hold-Out checkpoint, one per
z-source. The example makes the collapse of \S\ref{sec:modeb-fails}
visible at the surface level: classifier inversion produces a degenerate
token-repetition loop, whereas \textsc{Ref-Enc} and \textsc{Prior}
produce fluent restaurant-review text. Samples are verbatim, lowercased
as in the corpus, and truncated to their first \textasciitilde$30$ words.

\begin{table}[h]
\centering
\small
\begin{tabular}{@{}p{0.15\columnwidth}p{0.77\columnwidth}@{}}
\toprule
z-source & Generated text (target: \texttt{negative\_female\_bar\_past}) \\
\midrule
\textsc{Cls-Inv} & \emph{class class class Class Class Class Kids Class Class Sing Kid Class Kids Special Kids Special Children Class Kids Kids Special Special Kids Children\,\dots} \\
\addlinespace
\textsc{Ref-Enc} & adios everyone, this place is totally worth going back to. the owner, who was seated at the bar, told us he had come to check out and that he had some frien\,\dots \\
\addlinespace
\textsc{Prior} & was so disappointed in this place. i ordered the last meal i had, which was a very small portion of the pizza. the waiter was so nice and attentive\,\dots \\
\bottomrule
\end{tabular}
\caption{Representative generations per z-source on GPT-2-124M Fyelp
Hold-Out for target \texttt{negative\_female\_bar\_past}. Classifier
inversion degenerates into a repetition loop, while \textsc{Ref-Enc} and
\textsc{Prior} produce fluent restaurant-review text; only \textsc{Prior}
clearly reflects the negative target in this example.}
\label{tab:samples}
\end{table}

\section{LLM-as-judge qualitative evaluation}
\label{app:judge}

The headline numbers rest on automatic metrics (official classifier
accuracy, perplexity, Dist-$n$). Because one of those metrics --- the
official classifier --- has a documented default-prediction artefact on
degraded text (App.~\ref{app:singleton}), we add a small LLM-as-judge
study as an independent qualitative check.

\paragraph{Protocol.} From the GPT-2-124M Fyelp Hold-Out
\texttt{test\_seen} generations we drew $20$ target combinations (seed
$42$) and, for each, the generation under every z-source --- $60$ texts,
combination-matched across z-sources. An LLM judge (Claude Opus 4.7) rated each
text under a fixed rubric, applied per generation. The rubric has two
parts. \textbf{(a) Attribute match}: for each of the three
policy-controlled axes (sentiment, cuisine, tense; gender excluded per the
ethics scope of \S\ref{sec:limitations}), the judge scores $1$ if the text
exhibits the target value for that axis and $0$ otherwise --- sentiment by
the review's overall polarity, cuisine by the food/venue named, tense by
the dominant verb tense --- summed to a $0$--$3$ attribute-match score.
\textbf{(b) Fluency}: a single categorical label --- \emph{fluent} (a
coherent, on-domain review), \emph{degraded} (grammatical but incoherent or
off-domain), or \emph{token-salad} (repetition loop or word/character
soup). Under uniform random guessing the per-axis hit rates are $0.5$
(sentiment, binary), $0.2$ (cuisine, $5$-way), and $0.5$ (tense, binary),
so chance scores $\approx 1.2$ on attribute match. 

\begin{table}[h]
\centering
\small
\begin{tabular}{lccc}
\toprule
 & attr.\ match & \multicolumn{2}{c}{fluency ($n=20$)} \\
z-source & (mean\,/3) & fluent & degraded\,/\,salad \\
\midrule
\textsc{Cls-Inv}        & 0.80 & 0  & 11 / 9 \\
\textsc{Ref-Enc}        & 1.90 & 16 & 4 / 0 \\
\textbf{\textsc{Prior}} & \textbf{2.45} & \textbf{20} & 0 / 0 \\
\bottomrule
\end{tabular}
\caption{LLM-as-judge evaluation: $20$ combination-matched generations per
z-source, GPT-2-124M Fyelp Hold-Out \texttt{test\_seen}, judge = Claude
under a fixed rubric. Attribute match counts the three policy-controlled
axes; chance guessing $\approx 1.2/3$. The prior leads on both attribute
match and fluency, and classifier inversion yields no fluent review.}
\label{tab:judge}
\end{table}

\paragraph{Findings (Tab.~\ref{tab:judge}).} \textbf{(1) The prior satisfies the attributes:} it
scores $2.45/3$ on attribute match, against $1.90$ for \textsc{Ref-Enc}
and $0.80$ for \textsc{Cls-Inv} --- the last \emph{below} the $\approx
1.2/3$ chance baseline, because collapsed text often expresses no
attribute at all. \textbf{(2) The prior does not sacrifice naturalness:}
all $20$ prior generations are rated fluent, slightly ahead of
\textsc{Ref-Enc} ($16/20$) --- the conditional-mean code does not trade
fluency for control. \textbf{(3) The single-sample
idiosyncrasy of \textsc{Ref-Enc} is visible:} \textsc{Ref-Enc} trails the prior on attribute
match ($-0.55$) and fluency ($4/20$ degraded vs.\ $0$), because an
individual reference encoding carries the quirks of the chosen sentence ---
wrong tense, off-topic drift --- exactly the sample-specific noise that the
conditional mean removes (\S\ref{sec:analysis-prior}). \textbf{(4)
Classifier-inversion failure is not only repetition:} no \textsc{Cls-Inv}
generation is a fluent review; $9/20$ are token-salad (repetition loops,
word or character soup) and $11/20$ are degraded --- grammatical but
off-domain \emph{pretrain-mode} text (city news, TV-show, essay
fragments) carrying no attribute signal. The collapse is therefore both
degenerate looping and reversion to the pretraining prior of the backbone,
both consistent with an off-manifold control code
(\S\ref{sec:analysis-modeb}).

\section{Off-manifold distance of the inverted code}
\label{app:offmanifold}

Tab.~\ref{tab:offmanifold} reports the direct measurement behind
\S\ref{sec:analysis-modeb}: the distance of each inference-time control
code $\mathbf{z}^\star$ to the training-data code distribution of the encoder,
on the Fyelp Hold-Out checkpoints.

\begin{table}[h]
\centering
\small
\begin{tabular}{llcc}
\toprule
Backbone & $\mathbf{z}^\star$ source & Mahalanobis & kNN-dist \\
\midrule
\multirow{3}{*}{GPT-2 124M}
 & \textsc{Cls-Inv} & $3.34_{\pm0.06}$ & $17.6_{\pm0.5}$ \\
 & \textsc{Ref-Enc}  & $1.00_{\pm0.21}$ & $4.5_{\pm0.5}$ \\
 & \textbf{\textsc{Prior}} & $\mathbf{0.50}_{\pm0.05}$ & $\mathbf{3.2}_{\pm0.1}$ \\
\midrule
\multirow{3}{*}{LLaMA-3.2 1B}
 & \textsc{Cls-Inv} & $3.74_{\pm0.15}$ & $10.2_{\pm0.9}$ \\
 & \textsc{Ref-Enc}  & $1.00_{\pm0.20}$ & $2.4_{\pm0.2}$ \\
 & \textbf{\textsc{Prior}} & $\mathbf{0.60}_{\pm0.09}$ & $\mathbf{1.7}_{\pm0.1}$ \\
\bottomrule
\end{tabular}
\caption{Distance of each inference-time control code to the encoder's
training-data code distribution (Fyelp Hold-Out; mean$_{\pm\text{sd}}$
across the $39$ seen combinations). Diagonal-Gaussian Mahalanobis and
mean distance to $10$ nearest training neighbours. Classifier inversion
lands $3$--$7\times$ farther off-manifold than the prior or an encoded
reference, on both backbones, with the gap many times its
combination-level spread.}
\label{tab:offmanifold}
\end{table}

\section{Manifold-regulariser sweep}
\label{app:clsinvreg}

Tab.~\ref{tab:clsinv-reg} gives the full $\beta$ sweep behind the
regulariser ablation in \S\ref{sec:modeb-fails}: classifier inversion
re-run with a label-agnostic manifold regulariser on the per-axis
objective, using two penalty forms, two backbones, and three seeds.

\begin{table}[h]
\centering
\small
\setlength{\tabcolsep}{3pt}
\begin{tabular}{lcccc}
\toprule
 & \multicolumn{2}{c}{GPT-2-124M} & \multicolumn{2}{c}{LLaMA-3.2 1B} \\
\cmidrule(lr){2-3}\cmidrule(lr){4-5}
$\beta$ & Mahal. & shell & Mahal. & shell \\
\midrule
$0$    & $43.8_{\pm0.6}$ & $43.8_{\pm0.6}$ & $41.7_{\pm1.2}$ & $40.5_{\pm0.5}$ \\
$0.03$ & $55.7_{\pm1.3}$ & $51.5_{\pm0.6}$ & $49.8_{\pm0.3}$ & $43.9_{\pm0.8}$ \\
$0.1$  & $\mathbf{57.4}_{\pm1.2}$ & $\mathbf{52.2}_{\pm0.4}$ & $47.5_{\pm0.8}$ & $43.0_{\pm0.7}$ \\
$0.3$  & $54.0_{\pm0.8}$ & $51.7_{\pm1.3}$ & $\mathbf{52.8}_{\pm0.4}$ & $49.1_{\pm0.1}$ \\
$1$    & $50.5_{\pm0.3}$ & $50.7_{\pm1.6}$ & $48.5_{\pm1.2}$ & $\mathbf{49.6}_{\pm0.8}$ \\
$3$    & $48.7_{\pm1.1}$ & $50.8_{\pm1.5}$ & $48.1_{\pm1.0}$ & $52.0_{\pm5.1}$ \\
\bottomrule
\end{tabular}
\caption{Classifier inversion with a label-agnostic manifold regulariser
of weight $\beta$: 4-axis seen accuracy (\%), mean$_{\pm\text{sd}}$ over
3 seeds (42/7/11). Fyelp, official evaluator, fixed $312$-row stratified
\texttt{test\_seen} subset ($8$ reviews per seen combination); $\beta=0$
reproduces plain classifier inversion. \emph{Mahal.}:
$((\mathbf{z}-\boldsymbol{\mu})/\boldsymbol{\sigma})^2$, pulls toward the
per-axis centroid. \emph{shell}:
$\mathrm{relu}(((\mathbf{z}-\boldsymbol{\mu})/\boldsymbol{\sigma})^2-\rho^2)$,
$\rho=2$ --- zero inside the $\pm2\sigma$ ellipsoid, no centring pull.
Every $\beta>0$ lifts accuracy off chance on both backbones and both
penalties (bold: per-column peak); the $+8$ to $+14$\,pp lift exceeds the
seed noise ($\le1.6$\,pp on $23/24$ cells) by roughly $10\times$.}
\label{tab:clsinv-reg}
\end{table}

\paragraph{Full-test-set regularised inversion.} Tab.~\ref{tab:clsinv-reg}
sweeps $\beta$ on a stratified $312$-row subset; we also ran the
best Mahalanobis setting on the \emph{full} Hold-Out test sets. The lift
holds: 4-axis seen accuracy rises from $42.99$ to $57.52$ on GPT-2-124M
and from $41.47$ to $52.91$ on LLaMA-3.2 1B, matching the subset sweep.
The regularised optimum still trails the label prior --- by
$10.2$ and $28.8$\,pp on the seen split and $4.5$ and $23.5$\,pp on the
unseen split. A sharpened \emph{label-conditioned} variant that pulls
toward the per-(axis, target-label) marginal mean
(Eq.~\ref{eq:clsinv-reg-cond}, Tab.~\ref{tab:clsinv-reg-cond}) modestly
improves over the label-agnostic centroid on GPT-2 ($+3.2$\,pp) but is
tied on LLaMA, still trailing $g_\gamma$ by $7$ and $29$\,pp ---
the advantage of the prior is structural (per-combination, not per-axis),
not merely optimisation cost.

\paragraph{Perplexity and diversity along the sweep.} On the GPT-2-124M
Hold-Out subset the regulariser collapses perplexity from $131.6$
($\beta=0$, bare classifier inversion) to $18$--$23$ for every $\beta>0$,
while 4-axis accuracy peaks at $58.7\%$ ($\beta=0.1$) and then declines.
Distinct-2 is \emph{highest} at $\beta=0$ ($0.50$) and falls to
$\sim\!0.30$ as $\beta$ grows: the elevated diversity of bare classifier
inversion is the token-salad signature (random $n$-grams), not useful
variety, and must be read alongside the $131.6$ perplexity.

\paragraph{Label-conditioned variant.} A natural sharpening of the
label-agnostic penalty in Tab.~\ref{tab:clsinv-reg} replaces the
per-axis global mean $\boldsymbol{\mu}_a$ with the per-(axis,
target-label) \emph{conditional} mean, conditioned on the inference
target $c_a^\star$:
\begin{equation}
R_{\text{cond}}(\mathbf{z}_a) = \mathrm{mean}\!\Big(\!\big(
(\mathbf{z}_a - \boldsymbol{\mu}_a^{c_a^\star}) /
\boldsymbol{\sigma}_a^{c_a^\star}\big)^2\Big),
\label{eq:clsinv-reg-cond}
\end{equation}
where $(\boldsymbol{\mu}_a^v, \boldsymbol{\sigma}_a^v)$ are the per-axis
mean/std of the encoder code over training samples whose axis-$a$ label
is $v$. Conditioning the pull on the target label is a strict
sharpening of the label-agnostic centroid and the natural candidate to
close the gap to the prior. Tab.~\ref{tab:clsinv-reg-cond} reports the
same $\beta$ sweep as Tab.~\ref{tab:clsinv-reg} on the same $312$-row
stratified subset, single seed (seed~42).

\begin{table*}[t]
\centering
\small
\setlength{\tabcolsep}{6pt}
\begin{tabular}{lcc}
\toprule
$\beta$ & GPT-2-124M & LLaMA-3.2 1B \\
\midrule
$0.1$ & $\mathbf{60.74}$ & $\mathbf{52.73}$ \\
$0.3$ & $56.65$ & $48.31$ \\
$1.0$ & $53.93$ & $49.92$ \\
$3.0$ & $54.89$ & $49.36$ \\
\midrule
\multicolumn{3}{l}{\textit{Reference (from Tab.~\ref{tab:clsinv-reg}, 3-seed mean):}} \\
Mahal., best $\beta$ (label-agnostic) & $57.4$ & $52.8$ \\
prior $g_\gamma$ (\S\ref{sec:analysis-prior})  & $\sim\!67.7$ & $\sim\!81.7$ \\
\bottomrule
\end{tabular}
\caption{Label-conditioned manifold regulariser
(Eq.~\ref{eq:clsinv-reg-cond}), 4-axis seen accuracy (\%) on the same
312-row stratified Fyelp Hold-Out \texttt{test\_seen} subset as
Tab.~\ref{tab:clsinv-reg}, single seed (42). Bold: per-column peak.
The label-conditioned penalty modestly outperforms the label-agnostic
centroid on GPT-2-124M ($+3.2$\,pp at $\beta=0.1$) and roughly ties on
LLaMA-3.2 1B ($-0.1$\,pp), yet remains $7$ and $29$\,pp below
$g_\gamma$. The per-axis conditional marginal collapses the cross-axis
structure that $g_\gamma$ amortises directly per combination, so it
cannot reach the prior's per-combination joint pull target.}
\label{tab:clsinv-reg-cond}
\end{table*}

\paragraph{Why label-conditioning cannot close the gap.} The
label-agnostic penalty imposes only a manifold constraint --- ``stay
near the training distribution of the encoder'' --- and leaves the
directional choice to the cross-entropy loss against $\mathbf{W}_a$.
The label-conditioned penalty also specifies \emph{where on
the manifold to head}: toward the per-axis marginal mean
$\boldsymbol{\mu}_a^{c_a^\star}$. This direction averages over all
training combinations that share $c_a^\star$ on axis $a$, regardless
of the labels of the other axes. The post-hoc prior $g_\gamma$ avoids this
averaging by amortising \emph{per-combination} means directly; that
structural distinction (per-axis marginal $\to$ per-combination joint),
not merely per-query optimisation cost, is why the prior beats every
regularised inversion variant we tested.

\section{Bare cls-inv: inversion-side hyperparameter grid}
\label{app:clsinvgrid}

The bare-inversion collapse in \S\ref{sec:modeb-fails} is measured at
a single inversion-side configuration ($50$ Adam steps, lr $=0.1$,
initialised from $\mathbf{W}_a[c_a^\star,:]^\top$). To rule out the
alternative explanation that this single setting is merely
\emph{under-optimised}, we sweep
$(\text{lr}, \text{steps}) \in \{0.01, 0.1, 1.0\}\times\{50, 200, 1000\}$
on the same 312-row Fyelp Hold-Out \texttt{test\_seen} subset used in
App.~\ref{app:clsinvreg}, single seed (42), no regulariser
($\beta=0$), with all other settings identical to the main run
(Tab.~\ref{tab:clsinvgrid}). Chance is
$42.5\%$ (uniform over $4{\times}2{\times}5{\times}3$ axis labels with
the per-axis size mix the evaluator uses).

\begin{table}[h]
\centering\small
\setlength{\tabcolsep}{6pt}
\begin{tabular}{lccc}
\toprule
& \multicolumn{3}{c}{steps} \\
\cmidrule(lr){2-4}
lr & $50$ & $200$ & $1000$ \\
\midrule
\multicolumn{4}{l}{\textit{GPT-2-124M (paper default in bold):}} \\
$0.01$  & $42.53$ & $43.57$ & $42.48$ \\
$0.1$   & $\mathbf{43.63}$ & $41.91$ & $42.28$ \\
$1.0$   & $46.61$ & $45.74$ & $46.96$ \\
\midrule
\multicolumn{4}{l}{\textit{LLaMA-3.2 1B at steps$=1000$:}} \\
$0.01$ & --- & --- & $44.23$ \\
$0.1$  & --- & --- & $41.74$ \\
$1.0$  & --- & --- & $45.75$ \\
\bottomrule
\end{tabular}
\caption{Bare \textsc{Cls-Inv} accuracy (\%) on a $3\times3$
inversion-side grid (GPT-2-124M) plus three LLaMA-3.2 1B sanity cells,
Fyelp Hold-Out \texttt{test\_seen} 4-axis mean. All $12$ cells lie
within $\pm 5$\,pp of the $42.5\%$ chance line; the best cell (lr $=1.0$,
GPT-2) reaches $46.96\%$, still $10.5$\,pp below the cheapest manifold
regulariser (Mahal.\ label-agn., $57.52\%$, Tab.~\ref{tab:clsinv-reg})
and $20.7$\,pp below the post-hoc prior. The bare-inversion collapse is
therefore not an under-optimisation artefact of the paper's
default schedule.}
\label{tab:clsinvgrid}
\end{table}

\section{Density-modelling baseline: conditional normalising flow}
\label{app:flowprior}

The post-hoc prior $g_\gamma$ in the main text is a deterministic MLP
fitted to the per-combination encoder \emph{means}. The natural
sharper alternative --- and the closest competitor in the latent-CTG
literature \citep{gu2023prior} --- is a learned \emph{conditional
density} $p(\mathbf{z}\mid c)$. If the within-combination encoder
variance carries useful signal, modelling the full conditional
distribution should outperform predicting only its mean. The
denoiser interpretation in \S\ref{sec:analysis-prior} predicts the
opposite: removing within-combination noise is what the deterministic
mean delivers, so a density model should \emph{underperform} the MLP.

We fit an $8$-layer conditional RealNVP $p(\mathbf{z}\mid c)$ on the
frozen encoder's training $(\mathbf{z}, c)$ pairs (per-combination
one-hot labels), max-likelihood for $5000$ Adam steps at lr $=10^{-3}$,
$128$-unit affine-coupling MLPs. At inference, we sample
$\mathbf{z}\sim p(\mathbf{z}\mid c^\star)$ given the target labels. The
flow uses the same frozen encoder, the same generator checkpoint, and
the same external CompMCTG evaluator as every other row of
Tab.~\ref{tab:protocol-compare}.

\paragraph{Result.} On Fyelp Hold-Out \texttt{test\_seen}, the
flow-prior reaches $57.21\%$ (GPT-2-124M) and $69.95\%$ (LLaMA-3.2 1B)
on the official 4-axis evaluator --- a genuine working baseline, well
above bare \textsc{Cls-Inv} ($+14.2$/$+28.5$\,pp) and above the
label-agnostic Mahalanobis regulariser on LLaMA ($+17$\,pp). It
still trails the deterministic MLP $g_\gamma$ by
$\mathbf{10.5}$\,pp (GPT-2) and $\mathbf{11.8}$\,pp (LLaMA-1B). The
ordering is consistent with the denoiser hypothesis: the density
target $p(\mathbf{z}\mid c^\star)$ retains the within-combination
variance that the conditional-mean estimator $\mathbb{E}[\mathbf{z}\mid
c^\star]$ averages out, so sampling from the flow injects
sample-specific variation that the injector of the generator then has to
process. The implicit denoising of $g_\gamma$ is therefore not a
side-effect of architectural simplicity --- a strictly more
expressive density model with the same fitting data underperforms by
roughly $11$\,pp on each backbone.

\section{Nearest-seen-combo retrieval baseline}
\label{app:retrieval}

To check whether the learned prior $g_\gamma$ does more than retrieve the
closest observed combination, we compare it with a non-parametric
baseline. Given a target combination, the baseline returns the
per-combination mean encoder code of the nearest seen combination, using
Hamming distance in label-index space; a seen target retrieves itself.
Both methods are evaluated under the official CompMCTG evaluator on the
four Fyelp checkpoints (Tab.~\ref{tab:retrieval}).

\begin{table}[h]
\centering
\small
\begin{tabular}{lcccc}
\toprule
 & \multicolumn{2}{c}{seen} & \multicolumn{2}{c}{unseen} \\
\cmidrule(lr){2-3}\cmidrule(lr){4-5}
Run & retr. & prior & retr. & prior \\
\midrule
GPT-2 124M, HO    & 68.8 & 67.7 & 46.6 & \textbf{55.2} \\
GPT-2 124M, ACD   & 74.6 & 74.2 & 53.3 & \textbf{55.9} \\
LLaMA-3.2 1B, HO  & 83.0 & 81.7 & 66.9 & \textbf{76.4} \\
LLaMA-3.2 1B, ACD & 87.0 & 86.9 & 64.0 & \textbf{69.5} \\
\bottomrule
\end{tabular}
\caption{Nearest-seen-combination retrieval baseline versus the learned
label prior $g_\gamma$: 4-axis accuracy (\%) under the official CompMCTG
evaluator. On seen combinations, retrieval and the prior agree within
$1.3$\,pp, as both recover the observed per-combination mean. On every
unseen split, the learned prior wins by $+2.6$ to $+9.5$\,pp.}
\label{tab:retrieval}
\end{table}

On seen combinations, retrieval and the learned prior are nearly
equivalent, as expected. On unseen combinations, however, the prior wins
consistently. This shows that $g_\gamma$ is not merely a lookup table:
its learned label-to-code map performs useful compositional interpolation
where compositional generalisation is tested.

\section{Isolating within-combination denoising: the $k$-sample control}
\label{app:ksample}

\S\ref{sec:analysis-prior} attributes the advantage of the prior over a
single-sample reference encoding to conditional-mean denoising. We test
that directly with a non-parametric control. For each \texttt{test\_seen}
target combination we draw $k$ training references \emph{of that same
combination}, encode each, and average their concept codes; the averaged
code drives generation in place of $g_\gamma(\mathbf{c}^\star)$. As $k$
grows this is an increasingly low-variance estimate of the per-combination
mean $\mathbb{E}[E_\phi(\mathbf{x})\mid\mathbf{c}]$ --- exactly the
quantity that Eq.~\ref{eq:prior} fits. The control is defined on
\texttt{test\_seen} only: unseen combinations have no training references.
We sweep $k\in\{1,4,16,64\}$ with three independent reference draws each,
plus a single $k{=}256$ draw, on GPT-2-124M for both the Hold-Out and ACD
Fyelp checkpoints, scored by the official 4-axis evaluator. The $k{=}1$
row here is one training-pool reference scored on \texttt{test\_seen}; it
is therefore not the \textsc{Ref-Enc} row of Tab.~\ref{tab:fyelp}, which
encodes each test sentence's own text and is scored on
\texttt{test\_unseen}.

\begin{table}[h]
\centering
\small
\begin{tabular}{lcc}
\toprule
$k$ references & Hold-Out & ACD \\
\midrule
$1$   & $61.52_{\pm2.00}$ & $67.62_{\pm0.74}$ \\
$4$   & $62.70_{\pm0.31}$ & $69.38_{\pm0.43}$ \\
$16$  & $63.35_{\pm0.31}$ & $70.60_{\pm0.26}$ \\
$64$  & $63.65_{\pm0.26}$ & $70.07_{\pm0.16}$ \\
$256$ & $68.05$           & $74.45$           \\
\midrule
prior $g_\gamma$              & $67.67$ & $74.23$ \\
retrieval ($k\!\to\!\infty$)  & $68.8$  & $74.6$  \\
\bottomrule
\end{tabular}
\caption{The $k$-sample within-combination averaging control: 4-axis
\texttt{test\_seen} accuracy (\%) under the official evaluator,
mean$_{\pm\text{sd}}$ over three reference draws for $k\le64$. Averaging
more same-combination references denoises the code: accuracy rises with
$k$ and the cross-draw spread collapses. By $k{=}256$ the non-parametric
average reaches the learned prior $g_\gamma$ to within $0.4$\,pp on both
datasets.}
\label{tab:ksample}
\end{table}

Three things hold on both datasets (Tab.~\ref{tab:ksample}).
\textbf{(i)} Accuracy increases with $k$ --- a single reference is a noisy
code, and averaging denoises it. \textbf{(ii)} The cross-draw standard
deviation collapses, from $\pm2.0$ at $k{=}1$ to below $\pm0.3$ for
$k\ge64$ --- the $\propto\!1/\sqrt{k}$ signature of a conditional-mean
estimator. \textbf{(iii)} By $k{=}256$ the average reaches $68.05$
(Hold-Out) and $74.45$ (ACD), within $0.4$\,pp of the learned prior
$g_\gamma$. In the limit $k\to\infty$ the within-combination average is,
by construction, the full per-combination training mean; we verified
numerically that this mean is byte-identical (cosine $1.000$) to the
nearest-seen-combo code of the retrieval baseline
(App.~\ref{app:retrieval}), whose accuracy ($68.8$ / $74.6$) brackets
$g_\gamma$. The edge of the prior over a single-sample \textsc{Ref-Enc} is
thus confirmed to be within-combination denoising: averaging enough references
of a combination reproduces the accuracy of the prior.

\section{Concept-code additivity and the factorised prior}
\label{app:additivity}

This appendix supports the additivity analysis in
\S\ref{sec:prior-recovers}. We ask three questions: how additive the
frozen-encoder concept code is, whether the non-additive residual is
explained by label co-occurrence, and whether adding an explicit
interaction term improves the label prior.

\paragraph{The concept code is only partially additive.}
We fit an additive per-axis model to the frozen-encoder code,
\begin{equation}
\begin{aligned}
\hat{\mathbf{z}}
&= \boldsymbol{\mu} + \sum_{a=1}^{A} \Delta_a(y_a), \\
\Delta_a(v)
&= \mathbb{E}[\mathbf{z}\mid y_a=v] - \boldsymbol{\mu}.
\end{aligned}
\label{eq:additive-code}
\end{equation}
and report the fraction of variance explained on the Fyelp training pool
(Tab.~\ref{tab:additivity}). The additive model explains $31.7\%$ of the
variance at GPT-2 124M, $38.3\%$ at GPT-2-Medium 355M, and $47.6\%$ at
LLaMA-3.2 1B. Thus additivity increases with backbone scale, but even the
largest backbone leaves more than half of the concept-code variance
unexplained by per-axis offsets. The per-block breakdown is also
informative: the gender block is weakly axis-localised at every scale
($5$--$12\%$), consistent with gender being the weakest attribute signal
in the CompMCTG Fyelp results.

\begin{table*}[t]
\centering
\small
\begin{tabular}{lccccccccc}
\toprule
 & \multicolumn{5}{c}{Additive decomposition (\% variance explained)} & \multicolumn{4}{c}{Concept-code variance} \\
\cmidrule(lr){2-6}\cmidrule(lr){7-10}
Backbone & Full & sent. & cuis. & tense & gend. & intra-std & inter-std & ratio & prior MSE \\
\midrule
GPT-2 124M       & 31.7 & 45 & 23 & 10 & 5  & 0.42 & 0.29 & 0.68 & 0.20 \\
GPT-2-Med.\ 355M & 38.3 & 54 & 38 & 13 & 6  & 0.43 & 0.33 & 0.76 & 0.20 \\
LLaMA-3.2 1B     & 47.6 & 67 & 51 & 20 & 12 & 0.25 & 0.25 & 1.02 & 0.07 \\
\bottomrule
\end{tabular}
\caption{Structure of the frozen-encoder concept code $\mathbf{z}$ on the
Fyelp training pool. \emph{Additive decomposition}: percentage of
variance explained by an additive per-axis model
$\boldsymbol{\mu}+\sum_a\Delta_a(y_a)$, and by each concatenated
per-axis block alone. \emph{Concept-code variance}: mean per-dimension
within-combination and between-combination standard deviation, their
ratio, and the residual MSE of a factorised label prior. Additivity rises
with backbone scale, but within-combination variation is comparable to or
larger than between-combination variation, suggesting a substantial
sample-specific component in the encoder code.}
\label{tab:additivity}
\end{table*}

%
%

\begin{figure}[t]
\centering
\begin{tikzpicture}[scale=0.92, font=\small]
  \foreach \rname [count=\ri from 0] in {sentiment,gender,cuisine,tense} {
    \foreach \cname [count=\ci from 0] in {sentiment,gender,cuisine,tense} {
      \ifnum\ri=\ci
        \fill[black!9] (\ci,-\ri) rectangle ++(1,-1);
        \node at (\ci+0.5,-\ri-0.5) {$-$};
      \else
        \fill[blue!7] (\ci,-\ri) rectangle ++(1,-1);
        \node[font=\footnotesize] at (\ci+0.5,-\ri-0.5) {0.00};
      \fi
    }
  }
  \draw[gray!40] (0,0) grid (4,-4);
  \foreach \cname [count=\ci from 0] in {sentiment,gender,cuisine,tense}
    \node[rotate=38,anchor=south west,font=\footnotesize]
      at (\ci+0.32,0.06) {\cname};
  \foreach \rname [count=\ri from 0] in {sentiment,gender,cuisine,tense}
    \node[anchor=east,font=\footnotesize] at (-0.12,-\ri-0.5) {\rname};
\end{tikzpicture}

\caption{Inter-attribute statistical dependence on the full Fyelp corpus
($70$K records, all $40$ attribute combinations). Each off-diagonal cell
is the normalised mutual information ($\mathrm{NMI} \in [0,1]$) between two
attribute axes. \textbf{Every pair is exactly $0.00$} --- raw mutual
information and Cram\'er's $V$ are likewise exactly zero, and every
value-pair pointwise mutual information is $+0.000$. The four Fyelp axes
are \emph{independent by construction}: CompMCTG samples attribute
combinations uniformly. The non-additive residual of the concept code
(\S\ref{sec:additivity}) therefore cannot be attribute interaction --- it
is an encoder artefact.}
\label{fig:pmi}
\end{figure}

\paragraph{The residual is not explained by label co-occurrence.}
A natural interpretation of the non-additive residual is that the
attributes interact, e.g., that the code for \textit{negative}
$\times$ \textit{Mexican} is not a linear superposition of its parts. We
therefore measure statistical dependence between the four Fyelp axes on
the full corpus. All six axis pairs have zero mutual information,
normalised mutual information, and Cram\'er's $V$; value-pair PMI is also
$+0.000$ throughout (Fig.~\ref{fig:pmi}). This confirms that Fyelp does
not contain measurable label co-occurrence structure for the prior to
exploit. The remaining non-additivity is therefore better interpreted as
arising from the learned encoder representation, rather than from
dataset-level attribute dependence.

\paragraph{Most recoverable signal is already captured by the mean.}
The residual can contain two components. The first is deterministic
encoder entanglement: the encoder may mix otherwise independent axes in a
way that a richer label-conditioned prior could, in principle, recover.
The second is per-sample variation: two reviews with the same attribute
combination can encode to different $\mathbf{z}$ values because of surface
content, style, or wording. This component is not recoverable from labels
alone and is precisely the variation removed by the conditional-mean
prior.

The within-versus-between decomposition suggests that the second
component is substantial. On GPT-2-Medium, the mean per-dimension standard
deviation is $0.43$ within a fixed combination, compared with $0.33$
across per-combination means. The within-combination variance therefore
sets a noise floor for any deterministic label-only predictor. A
factorised prior fitted to the same checkpoint reaches residual MSE
$0.20$, close to this floor; Tab.~\ref{tab:additivity} shows the same
pattern across backbones.

\paragraph{Explicit interaction terms do not improve the prior.}
We also test whether a richer factorised prior helps in practice: a
per-axis additive base plus a rank-24 interaction term. It matches the
plain prior within noise on GPT-2-Medium (seen $76.4$ vs.\ $78.5\%$,
unseen $57.7$ vs.\ $59.1\%$ on Fyelp ACD) and performs worse on
LLaMA-3.2 1B (seen $83.0$ vs.\ $86.9\%$, unseen $60.2$ vs.\ $69.5\%$),
with a larger seen-to-unseen drop. We therefore retain the plain additive
prior. The result should be read conservatively: in this benchmark, the
extra interaction capacity does not provide reliable gains and may
overfit seen combinations.

\section{Training schedule, loss-weight policy, and comparability policy}
\label{app:training}

\paragraph{Four-phase loss-weight schedule.}
Training follows the objective in Eq.~\ref{eq:total}. Let
$\tau_1$ denote the number of LM warm-up epochs, with default
$\tau_1=10$ for natural-language datasets and $\tau_1=5$ for the
synthetic task. All main runs train for $25$ epochs.

\begin{table*}[t]
\centering
\small
\begin{tabular}{lccc}
\toprule
Phase & Range & $\lambda_{\mathrm{c}}$ & $\lambda_{\mathrm{o}}$ \\
\midrule
1 (LM warm-up) & $\tau < \tau_1$ & $0$ & $0$ \\
2 (concept ramp) & $\tau_1 \le \tau < \tau_1 + 5$ & $\to 1$\textsuperscript{$\dag$} & $0$ \\
3 (orthogonality ramp) & $\tau_1 + 5 \le \tau < \tau_1 + 10$ & $1$ & $\to 0.1$\textsuperscript{$\dag$} \\
4 (full objective) & $\tau \ge \tau_1 + 10$ & $1$ & $0.1$ \\
\bottomrule
\end{tabular}
\caption{Four-phase loss-weight schedule for Eq.~\ref{eq:total}.
\textsuperscript{$\dag$}Linear ramp from $0$ over $5$ epochs. In phase 4,
the concept and orthogonality weights are fixed at
$\lambda_{\mathrm{c}}=1$ and $\lambda_{\mathrm{o}}=0.1$, respectively.}
\label{tab:schedule}
\end{table*}

\paragraph{Comparability policy.}
To keep the protocol comparisons interpretable, we use the following
policies throughout the main results.

\begin{description}
\setlength{\itemsep}{2pt}
\item[P1: Synthetic-task setting.]
All synthetic-task results use the 4-axis configuration
(purpose $\times$ rhetoric $\times$ structure $\times$ style), excluding
the topic axis.

\item[P2: Fixed evaluation seed.]
Training seeds vary only in designated multi-seed runs; the evaluation
seed is fixed at $42$ for sampling, prior-MLP initialisation, and
diagnostic pair selection.

\item[P3: Three z-source protocols.]
We report classifier inversion (\textsc{Cls-Inv},
Eq.~\ref{eq:clsinv}), single-sample reference encoding
(\textsc{Ref-Enc}), and the label-conditioned prior (\textsc{Prior},
Eq.~\ref{eq:prior}). \textsc{Ref-Enc} is a diagnostic, not an upper bound.

\item[P4: One stable recipe per backbone.]
Each backbone uses one validation-selected stable recipe, with no
within-row recipe mixing. Four main backbones use the AdaLN injector with
$\mathtt{inject\_every\_n}=1$; Qwen-2.5 1.5B uses the additive injector
with $g_{\mathrm{init}}=-3$ and $\mathtt{inject\_every\_n}=2$ after AdaLN
proved unstable on that backbone.

\item[P5: Unseen split as primary.]
Headline results use \texttt{test\_unseen}. We report
\texttt{test\_seen} when analysing seen--unseen gaps, matched-baseline
comparisons, or classifier-default artefacts.

\item[P6: Precision recipe fixed by backbone.]
GPT-2 124M and Qwen-2.5 1.5B train in fp32. Qwen-2.5 0.5B, LLaMA-3.2 1B,
and LLaMA-3.2 3B use a bf16 backbone with fp32 small modules, including
encoder MLPs, classifier heads, injector, and prior. App.~\ref{app:dtype}
reports the precision ablation.

\item[P7: Phase-4 loss weights fixed.]
All main runs use $\lambda_{\mathrm{c}}=1$ and
$\lambda_{\mathrm{o}}=0.1$ in phase 4, as shown in
Tab.~\ref{tab:schedule}. These constants are held fixed across backbones
and seeds.
\end{description}

\paragraph{Training hyperparameters.}
Unless otherwise specified, all main runs use AdamW
($\beta_1=0.9$, $\beta_2=0.999$), learning rate $5\times10^{-5}$ for
non-LoRA parameters and $2\times10^{-5}$ for LoRA adapters, cosine decay,
gradient clipping at norm $1.0$, and $25$ training epochs. Batch size is
$16$ for backbones up to $1.5$B and $4$ for LLaMA-3 8B due to memory
constraints. The per-axis concept dimension is $d_c=32$, with $A=4$ for
4-axis tasks and $A=1$ for YelpP.

All generators use LoRA rank $r=8$ and $\alpha=16$, targeting
attention-projection modules: \texttt{c\_attn} for GPT-2 and
\texttt{q\_proj}, \texttt{v\_proj} for Qwen-2.5 and LLaMA-3.2. The
label-prior MLP $g_\gamma$ has one hidden layer with $128$ GELU units and
is fitted post-hoc for $1000$ Adam steps with learning rate $10^{-3}$ and
batch size $64$. Encoding the training pool and fitting the prior takes
under $30$ seconds per checkpoint on a single GPU.

\section{Precision-recipe (dtype) ablation}
\label{app:dtype}

The precision recipe in P6 uses a bf16 backbone with fp32 small modules:
the concept-encoder MLPs, classifier heads, injector projections, and
prior MLP. We include a dtype ablation because early full-bf16 runs showed
substantially lower absolute accuracy on Qwen-0.5B and LLaMA-1B. Holding
the backbone in bf16 while casting the small modules to fp32 recovered
this loss. Tab.~\ref{tab:dtype} reports both recipes for the backbones
where both runs are available.

\begin{table*}[t]
\centering
\small
\begin{tabular}{lcc}
\toprule
& bf16-everywhere & bf16 + fp32 small \\
\midrule
\multicolumn{3}{l}{\textit{Synthetic 4-axis Acc$_{\text{4ax,unseen}}$ (mean)}} \\
LLaMA-3.2 1B (ref-enc)  & $0.655 \pm 0.098$ & $\mathbf{0.791 \pm 0.021}$ \\
LLaMA-3.2 1B (prior)   & $0.712 \pm 0.098$ & $\mathbf{0.840 \pm 0.026}$ \\
\quad $\Delta$ prior $-$ ref-enc & $+5.7 \pm 0.8$ & $\mathbf{+4.86 \pm 0.96}$ \\
\midrule
Qwen-2.5 0.5B (ref-enc) & $0.424$ & $\mathbf{0.603}$ \\
Qwen-2.5 0.5B (prior)  & $0.497$ & $\mathbf{0.679}$ \\
\quad $\Delta$ prior $-$ ref-enc & $+7.32$ & $\mathbf{+7.55}$ \\
\midrule
LLaMA-3.2 3B (ref-enc)  & $\approx 0.66$ & $\mathbf{0.857}$ \\
LLaMA-3.2 3B (prior)   & $\approx 0.71$ & $\mathbf{0.907}$ \\
\quad $\Delta$ prior $-$ ref-enc & $\approx +5.6$ & $\mathbf{+4.93}$ \\
\bottomrule
\end{tabular}
\caption{Precision-recipe ablation: bf16-everywhere vs.\ mixed precision
(bf16 backbone, fp32 small modules), four-axis synthetic, \texttt{test\_unseen}.
Absolute ref-enc/prior accuracies depend on dtype recipe by
$\sim \!\!+18$\,pp; the $\Delta$ \emph{magnitude} between prior and
ref-enc stays within $\pm 1.0$\,pp on all three backbones we measured.
LLaMA-1B numbers are 5-seed means (phase1c bf16-everywhere vs.\ phase1d
fp32-small modules); Qwen-0.5B and LLaMA-3B are single seed=42 (R5/R5d
and R7/R7d ablation pairs); LLaMA-3B bf16-everywhere absolute numbers
are approximate as the original bf16 run was not multi-seeded.}
\label{tab:dtype}
\end{table*}

\paragraph{Interpretation.} The dtype recipe shifts absolute \textsc{Ref-Enc} and \textsc{Prior}
accuracies in the same direction, but leaves the prior-over-\textsc{Ref-Enc}
gap stable. We therefore treat the dtype choice as an implementation
stability issue rather than a confound for the protocol comparison. The
most plausible source is bf16 rounding in the small MLPs and classifier
heads that operate on frozen-backbone activations; the LoRA-adapted
backbone itself appears to tolerate bf16. GPT-2 124M and Qwen-2.5 1.5B
were trained in full fp32 and are not included in this ablation because no
bf16 counterpart was run.

\section{Singleton-combo classifier-default artefact (Fyelp \texttt{test\_unseen}, \textsc{Cls-Inv})}
\label{app:singleton}

This appendix documents the singleton-combo classifier-default artefact
referenced in Tab.~\ref{tab:fyelp} footnote~$\dag$ and \S\ref{sec:setup}.

\paragraph{Symptom (Tab.~\ref{tab:singleton}).} The Fyelp Hold-Out
idx=$-0$ \texttt{test\_unseen} split contains a single attribute
combination, \texttt{negative\_male\_american\_present}. Classifier
inversion on LLaMA-3.2 1B produces token-salad output (PPL $1181 \pm
850$, $27\times$ ref-enc), yet the CompMCTG official RoBERTa-large
classifier reports an inflated 4-axis mean of $64.16 \pm 14.25\%$ ($n=5$). On the seen split
(39 distinct combinations), the same checkpoints score
$41.27 \pm 1.56\%$ --- within $1.3$\,pp of the random baseline $42.5\%$
and with $9\times$ smaller cross-seed variance.

\paragraph{Mechanism.} On degraded text, the official classifier tends to fall back to
dataset-prior predictions: \texttt{negative} for sentiment,
\texttt{male} for gender, \texttt{present} for tense, and a
non-\texttt{american} cuisine class. Three of these defaults match the
singleton unseen target, while the cuisine default does not. The result is
an artificially high 4-axis mean on a one-combination evaluation set. The
same pattern appears on other backbones: Qwen-2.5 1.5B shows high gender
and tense but near-chance cuisine on \texttt{test\_unseen}, and GPT-2 124M
shows the same gender/cuisine imbalance.

\begin{table*}[t]
\centering
\small
\begin{tabular}{lccc}
\toprule
Axis & target & seen-split prior & cls-inv unseen \\
\midrule
sentiment & negative & $\sim\!50\%$ & $83.79 \pm 27.62\%$ \\
gender    & male     & $\sim\!48\%$ & $\mathbf{100.0 \pm 0.0\%}$ \\
cuisine   & american & $\sim\!20\%$ & $5.15 \pm 6.58\%$\,\textsuperscript{$\ddag$} \\
tense     & present  & $\sim\!50\%$ & $67.70 \pm 37.37\%$ \\
\midrule
4-axis mean & --- & $\sim\!42\%$ chance & $64.16 \pm 14.25\%$ \\
\bottomrule
\end{tabular}
\caption{Per-axis breakdown of the singleton-combo artefact (LLaMA-1B
Fyelp \texttt{test\_unseen} \textsc{Cls-Inv}, $n=5$). \textsuperscript{$\ddag$}~cuisine
is the only axis whose classifier default ($\sim\!1/5$) does \emph{not}
match the singleton target \texttt{american}; the resulting cell is
below random and confirms the classifier defaults are not informative on
token-salad input.}
\label{tab:singleton}
\end{table*}

\paragraph{Why this is not the protocol succeeding.} Two independent
diagnostics confirm classifier inversion has collapsed even on
\texttt{test\_unseen} despite the inflated 4-axis number:
(i) the per-axis breakdown above does not look like a
concept-conditioned generation pattern (gender is locked at $100\%$ with
zero variance; cuisine is below random; sentiment and tense show
$> 25$\,pp variance across seeds);
(ii) the corresponding PPL is $1181 \pm 850$, roughly $27\times$
the ref-enc's $43.08 \pm 2.27$, consistent with the activation-level
over-shoot signature of §\ref{sec:analysis-modeb} ($\mathrm{rel\_mod}$ amplifying from
$3.3$ to $85$ across the $16$-layer LLaMA-1B stack).

Both diagnostics agree with the seen-split number ($41.27 \pm 1.56\%$,
within $0.4$\,pp of chance): the protocol genuinely collapses; the
\texttt{test\_unseen} 4-axis mean is an artefact of the
single-combination evaluation set interacting with the dataset-prior
default predictions of the classifier on degraded text. We therefore
report the seen-split number as the headline collapse evidence in
\S\ref{sec:modeb-fails} and use the \texttt{test\_unseen} \textsc{Cls-Inv}
numbers only as illustrations of the artefact.

\section{Artefact-free ACD comparison of prior and \textsc{Cls-Inv}}
\label{app:acdclsinv}

The within-split prior--\textsc{Cls-Inv} comparison on Fyelp Hold-Out
idx=$-0$ \texttt{test\_unseen} is affected by the singleton-combination
classifier-default artefact described in App.~\ref{app:singleton}. In
contrast, ACD idx=$-0$ holds out half of all attribute combinations, so
its \texttt{test\_unseen} split provides an artefact-free within-split
comparison (Tab.~\ref{tab:acdclsinv}).

\begin{table}[h]
\centering
\small
\begin{tabular}{lccc}
\toprule
Backbone & \textsc{Cls-Inv} & \textsc{Prior} & lift \\
\midrule
GPT-2 124M     & $48.47$ & $55.91$ & $+7.44$ \\
LLaMA-3.2 1B   & $50.75$ & $69.53$ & $+18.78$ \\
Qwen-2.5 1.5B  & $52.82$ & $60.48$ & $+7.66$ \\
\bottomrule
\end{tabular}
\caption{Artefact-free prior--\textsc{Cls-Inv} comparison on CompMCTG
Fyelp ACD \texttt{test\_unseen}, evaluated by the official RoBERTa-large
classifier suite. ACD holds out half of all attribute combinations, so the
evaluation does not reduce to a singleton target. The prior improves over
\textsc{Cls-Inv} by $+7.4$ to $+18.8$\,pp on matched checkpoints.}
\label{tab:acdclsinv}
\end{table}

This within-split comparison confirms the protocol ordering in
Tab.~\ref{tab:fyelp}: on the same checkpoints and without the singleton
artefact, the prior consistently recovers more compositional control than
\textsc{Cls-Inv}.

\section{Concept-swap fidelity}
\label{app:swap}

We use a concept-swap diagnostic to test whether the learned code
$\mathbf{z}$ is factored across axes or only coherent at the full
combination level. For two test samples $(u,v)$ that differ on every
axis, and for an axis $a$, we form
\[
\mathbf{z}_{\mathrm{swap}}^a =
[\mathbf{z}_v^{(0)},\dots,\mathbf{z}_u^{(a)},\dots,
\mathbf{z}_v^{(A-1)}].
\]
If the code is axis-factored, generation from
$\mathbf{z}_{\mathrm{swap}}^a$ should realise the axis-$a$ label of $u$
and the labels of $v$ on all other axes. We report mean joint fidelity
under the loose taxonomy: HIGH $>0.40$, PARTIAL $0.10$--$0.40$, and
ENT $<0.10$.

\begin{table}[h]
\centering
\small
\begin{tabular}{lcc}
\toprule
Backbone & Mean joint & Grade \\
\midrule
\multicolumn{3}{l}{\textit{Synthetic 4-axis, marginal-independent attributes}} \\
GPT-2 (R4)            & 0.028 & ENT \\
Qwen-0.5B (R5d)       & 0.037 & ENT \\
Qwen-1.5B (R2)        & 0.004 & ENT \\
LLaMA-1B ($n{=}5$)    & \textbf{0.145 $\pm$ 0.019} & \textbf{PARTIAL} \\
LLaMA-3B (R7d)        & \textbf{0.268} & \textbf{PARTIAL} \\
\midrule
\multicolumn{3}{l}{\textit{Fyelp natural data, seed=42}} \\
LLaMA-1B Fyelp        & 0.063 & ENT \\
Qwen-1.5B Fyelp       & 0.047 & ENT \\
\bottomrule
\end{tabular}
\caption{Concept-swap fidelity. Bins are heuristic: HIGH $>0.40$,
PARTIAL $0.10$--$0.40$, and ENT $<0.10$. On the synthetic task, LLaMA
shows partial axis-wise fidelity while GPT-2 and Qwen remain entangled.
On Fyelp, the same pattern does not transfer: both tested natural-data
checkpoints remain entangled under this diagnostic.}
\label{tab:swap}
\end{table}

The synthetic task shows a family-level pattern: LLaMA reaches the
PARTIAL range, increasing from $0.145$ at 1B to $0.268$ at 3B, whereas
GPT-2 and Qwen remain entangled. On Fyelp, however, this pattern does not
transfer; both tested natural-data checkpoints remain below the
entanglement threshold. Although the attribute
labels are marginally independent by construction
(\S\ref{sec:additivity}), their textual realisations can still be
entangled in the encoder representation. We therefore treat concept-swap
fidelity as a supplementary diagnostic and make no interpretability claim
from it in the main text.

\end{document}